\documentclass[11pt,bezier,amstex]{article}

\usepackage{graphicx,verbatim}
\usepackage[rflt]{floatflt}
\usepackage{epsfig,times,subfigure,graphicx}
\usepackage{amsmath,amssymb,amsopn,amsbsy,algorithm,algorithmic,theorem,float,bbm,bm,enumerate,color,multirow}
\usepackage{fullpage,setspace}
\usepackage{rotating}
\usepackage{array}
\usepackage[unicode]{hyperref}
\usepackage{url}
\usepackage{url}
\usepackage{chngcntr}
\usepackage{nameref}
\usepackage{zref-xr}
\usepackage{authblk}

\zxrsetup{toltxlabel}

\newcommand{\beq}{\begin{equation}}
\newcommand{\eeq}{\end{equation}}

\newcommand\E{\mathbb{E}}
\newcommand\I{\mathbb{I}}

\renewcommand\P{\mathbb{P}}
\newcommand\R{\mathbb{R}}

\newcommand{\g}{\mathbf{g}}

\renewcommand{\u}{\mathbf{u}}

\newcommand{\w}{\mathbf{w}}
\newcommand{\x}{\mathbf{x}}
\newcommand{\y}{\mathbf{y}}

\newcommand{\cF}{{\cal F}}

\newcommand{\cA}{{\cal A}}
\newcommand{\cB}{{\cal B}}
\newcommand{\cG}{{\cal G}}
\newcommand{\bgg}{\mathbf{g}}

\newcommand{\vertiii}[1]{{\left\vert\kern-0.25ex\left\vert\kern-0.25ex\left\vert #1
    \right\vert\kern-0.25ex\right\vert\kern-0.25ex\right\vert}}

\newcommand{\m}{\boldsymbol{\mu}}

\newcommand{\bxi}{\boldsymbol{\xi}}

\DeclareMathOperator{\argmax}{argmax}
\DeclareMathOperator{\argmin}{argmin}

\DeclareMathOperator{\tr}{Tr}

\newcounter{exampleI}
{\theorembodyfont{\rmfamily} \theoremstyle{plain} }

\newcounter{exampleII}
{\theorembodyfont{\rmfamily} \theoremstyle{plain} }

\newcounter{exampleIII}
{\theorembodyfont{\rmfamily} \theoremstyle{plain} }

{\theorembodyfont{\rmfamily} }
{\theorembodyfont{\rmfamily} \newtheorem{defn}{Definition}}
\newtheorem{theo}{Theorem}
\newtheorem{prop}{Proposition}
\newtheorem{lemm}{Lemma}

\newcommand{\proof}{\noindent{\itshape Proof:}\hspace*{1em}}

\newcommand{\qed}{\nolinebreak[1]~~~\hspace*{\fill} \rule{5pt}{5pt}\vspace*{\parskip}\vspace*{1ex}}

\newcommand {\commentout}[1] {}

\def\ints{{{\rm Z} \kern -.35em {\rm Z} }}  % ints
\def\smallints{{{\rm Z} \kern -.3em {\rm Z} }}  % small ints
\def\pints{{{\rm I} \kern -.15em {\rm N} }}      % pints
\newcommand{\reals}{\mathbb R}

\def\cplx{{{\rm I} \kern -.45em {\rm C} }}       % complex
\def\l2{\rm {\mathcal L}^{2}(\reals)}            % l2

\newtheorem{nad}{Notation and Definitions}[section]

\newtheorem{corollary}{Corollary}

\newcommand{\be}{\begin{eqnarray}}
\newcommand{\ee}{\end{eqnarray}}
\newcommand{\bea}{\begin{eqnarray}}
\newcommand{\eea}{\end{eqnarray}}
\newcommand{\beaa}{\begin{eqnarray*}}
\newcommand{\eeaa}{\end{eqnarray*}}
\newcommand{\bnad}{\begin{nad}}
\newcommand{\enad}{\end{nad}}

\newcommand{\eps}{\varepsilon}

\newcommand{\calO}{{\cal O}}

\usepackage[suppress]{color-edits}
\addauthor{ab}{magenta}
\addauthor{sw}{blue}
\addauthor{vs}{red}
\addauthor{qg}{cyan}

\newcommand{\qg}[1] {{\bf (Qilong says: #1)}}

\title{Random Quadratic Forms with Dependence:\\
Applications to Restricted Isometry and Beyond}

\author{Arindam Banerjee}
\author{~~~Qilong Gu}
\author{~~~Vidyashankar Sivakumar}
\author{~~~Zhiwei Steven Wu}
\affil{Department of Computer Science \& Engineering\\ 
University of Minnesota, Twin Cities\\ 
Minneapolis, MN 55455, USA}

\begin{document}

\maketitle

\begin{abstract}
Several important families of computational and statistical results in machine learning and randomized algorithms rely on uniform bounds on quadratic forms of random vectors or matrices. Such results include the Johnson-Lindenstrauss (J-L) Lemma, the Restricted Isometry Property (RIP), randomized sketching algorithms, and approximate linear algebra. The existing results critically depend on statistical independence, e.g., independent entries for random vectors, independent rows for random matrices, etc., which prevent their usage in dependent or adaptive modeling settings. In this paper, we show that such independence is in fact not needed for such results which continue to hold
under fairly general dependence structures. In particular, we present uniform bounds on random quadratic forms of stochastic processes which are conditionally independent and sub-Gaussian given another (latent) process. Our setup allows general dependencies of the stochastic process on the history of the latent process and the latent process to be influenced by realizations of the stochastic process. The results are thus applicable to adaptive modeling settings and also allows for sequential design of random vectors and matrices. We also discuss stochastic process based forms of  J-L, RIP, and sketching, to illustrate the generality of the results. 

%At a high level, similar to the generalization that Azuma-Hoeffding provides on the basic Hoeffding bound, our result provides a similar generalization of existing results on supremum of quadratic forms of independent random variables to MDSs. 
\end{abstract}

%\abcomment{this is a test}

\section{Introduction}
\label{sec:intro}

Over the past couple of decades, a set of key developments in statistical machine learning and randomized algorithms have been relying on uniform large deviation bounds on quadratic forms involving random vectors or matrices. The Restricted Isometry Property (RIP) is a well-known and widely studied result of this type, which has had a major impact in high-dimensional statistics \cite{nrwy12,bcfs14,vers14,vers18}. The Johnson-Lindenstrauss (J-L) Lemma is another well known result of this type, which has led to major statistical and algorithmic advances in the context of random projections \cite{joli84,aich06,inmo98}. Similar substantial developments have been made in several other contexts, including sketching algorithms based on random matrices \cite{wood14,kane14}, advances in approximate linear algebra \cite{maho11,drkm06}, among others.

Such existing developments in one way or another rely on uniform bounds on quadratic forms of random vectors or matrices.
Let $\cA$ be a set of $(m \times n)$ matrices and $\bxi \in \R^n$ be a sub-Gaussian random vector~\cite{vers14,vers18}. The existing results stem from large deviation bounds of the following random variable~\cite{krmr14}:
\begin{equation}
    C_{\cA}(\bxi) = \sup_{A \in \cA} \left| \| A \bxi\|_2^2 - \E \| A \bxi\|_2^2 \right|~.
    \label{eq:cixi0}
\end{equation}
Results such as RIP and J-L can then be obtained in a straightforward manner (see Section~\ref{sec:app} for details) from such bounds by converting the matrix $A$ into a vector $\theta = \text{vec}(A)$ and converting $\bxi$ into a suitable random matrix $X$ to get bounds on  
\begin{equation}
    C_{\Theta}(X) = \sup_{\theta \in \Theta} \left| \| X \theta \|_2^2 - \E \| X \theta \|_2^2 \right|~,
\end{equation}
where $\Theta = \{ \text{vec}(A) | A \in \cA\}$.
Results on other domains such as sketching \cite{wood14,kane14} and approximate linear algebra \cite{maho11,drkm06} can be similarly obtained.
Further, note that such bounds are considerably more general than the popular Hanson-Wright inequality \cite{ruve13,hawr71} for quadratic forms of random vectors, which focus on a fixed matrix $A$ instead of a uniform bound over a set $\cA$.

The key assumption in all existing results is that the entries $\xi_j$ of $\bxi$ need to be {\em statistically independent}. Such independence assumption shows up as element-wise independence of the random vector $\bxi$ in quadratic forms like $C_{\cA}(\bxi)$ and row-wise or element-wise independence of the random matrix $X$ in quadratic forms like $C_{\Theta}(X)$.
Existing analysis techniques, typically based on advanced tools from empirical processes \cite{vers14,ledo13}, rely on such independence to get uniform large deviation bounds.

In this paper, we consider a generalization of such existing results by allowing for statistical dependence in $\bxi$. 
In particular, we assume $\bxi = \{\xi_j\}$ to be a stochastic process where the marginal random variables $\xi_j$ are conditionally independent and sub-Gaussian given some other (latent) process $F=\{ F_j\}$. While hidden Markov models (HMMs)~\cite{barb12} are a simple example of such a setup, with $F$ being the latent variable sequence and $\bxi$ being the observations, our setup described in detail in Section~\ref{sec:setup} allows for far more complex dependencies, and allows for many different types of graphical models connecting $\bxi$ and $F$. For example, the setup allows graphical models where $\xi_j$ can have unrestricted statistical dependence on the full history $F_{1:j}$; further, the setup allows graphical models where $\xi_j$ can have unrestricted statistical dependence on the full history $F_{1:(j-1)}$, and $F_j$ has unrestricted statistical (or deterministic) dependence on the full history $F_{1:(j-1)}$ as well as $\xi_j$. The latter graphical model can in fact be considered adaptive since the realization of $\xi_j$ affects $F_j$ and in turn future $F_k, k > j$. In Section~\ref{sec:setup} we discuss two key conditions such graphical models need to satisfy and give a set of concrete examples of graphical models which satisfy the conditions illustrating the flexibility of the setup. Our main result is to establish a uniform large deviation bound for $C_{\cA}(\bxi)$ in \eqref{eq:cixi0} where $\bxi$ is any stochastic process following the setup outlined in Section~\ref{sec:setup}.

% Consider the factorization of a general joint distribution on $\bxi$, given by $p(\xi_1,\ldots,\xi_n) = p(\xi_1) p(\xi_2|\xi_1) \ldots p(\xi_n|\xi_1,\ldots,\xi_{n-1})$. Our new results (Section~\ref{sec:result}) only require the conditional distributions $p(\xi_j|\xi_1,\ldots,\xi_{j-1}), j=1,\ldots,n$ to be zero-mean sub-Gaussian. As long as this simple condition is satisfied, 
% the conditional distribution $\xi_j|\xi_1,\ldots,\xi_{j-1}$ can have arbitrary dependence on the history $\{\xi_1,\ldots,\xi_{j-1}\}$. More formally, we assume $\xi_j|\xi_1,\ldots,\xi_{j-1}$ to be a sub-Gaussian martingale difference sequence (MDS). The assumption is a substantial generalization of existing results which require $\bxi$ to be element-wise zero-mean and independent.

There are two broad types of implications of our results allowing for statistical dependence in random quadratic forms (Section~\ref{sec:app}). First, there are several emerging domains where data collection, modeling and estimation take place adaptively, including bandits learning, active learning, and time-series analysis \cite{auer03,sett12,lutk05}. The dependence in such adaptive settings is hard to handle, and existing analysis for specific cases goes to great lengths to work with or around such dependence~\cite{NTTZ18,DMST18,NR18}. The general tool we provide for such settings has the potential to simplify and generalize results in adaptive data collection, e.g., our results are applicable to the smoothed analysis of contextual linear bandits considered in \cite{kmrw18}. \abdelete{A critical component of their result depends on showing non-asymptotic RIP type results on design matrices whose rows are sub-Gaussian and generated sequentially with every row having dependence on previously observed rows. The current analysis has to design a workaround to use results from \cite{trop12} which requires the rows to have bounded $\ell_2$ norm. In contrast, application of our results does away with the $\ell_2$ norm bounded assumption yielding stronger results with much simpler analysis.} \vscomment{Check this. Also need to decide if we want to keep this based on the discussion in Section 4}  \abcomment{will be great to add 1-2 concrete lines on CLB (contextual linear bandits), say based on the discussion we (will) have in Section~\ref{sec:app}.} Second, since our results allow for sequential construction of random vectors and matrices adaptively, by considering realized elements or rows so far, randomized algorithmic approaches such as J-L and sketching would arguably be able to take advantage of such extra flexibility possibly leading to adaptive and more computationally efficient algorithms.  In Section~\ref{sec:app}, we illustrate how results such as J-L, RIP, and bandits would look like by allowing for dependence in the random vectors or matrices.  

The technical analysis for our main result is \swdelete{inspired by and is a generalization of} \swedit{a significant generalization of} prior analysis on tail behavior of chaos processes \cite{argi93,krmr14,tala14} for random vectors with i.i.d.~elements. To construct a uniform bound on $C_{\cA}(\bxi)$ in \eqref{eq:cixi0} for a stochastic process $\bxi$ with statistically dependent entries, we decompose the analysis into two parts, respectively depending on the off-diagonal terms and the diagonal terms of $A^T A$ from $\| A \bxi\|^2 = \bxi^T A^T A \bxi$. 
Our analysis for the off-diagonal terms
is based on two key tools: decoupling~\cite{pegi99} and generic chaining~\cite{tala14}, both with suitable generalizations from i.i.d.~counter-parts to stochastic processes $\bxi$. For decoupling, we present a new result on  decoupling of quadratic forms of sub-Gaussian stochastic processes $\bxi$ satisfying the conditions of our setup. Our result generalizes the classical decoupling result for vectors with i.i.d. entries~\cite{pegi99,krmr14}. 
%Note that such a decoupling result generalizes the corresponding result for random vectors with i.i.d.~entries based on an independent copy of the random vector \cite{krmr14,pegi99}.  
For generic chaining, we develop new results of interest in our context as well as generalize certain existing results for i.i.d.~random vectors to stochastic processes. While generic chaining, as a technique, does not need or rely on statistical independence~\cite{tala14}, an execution of the chaining argument does rely on an atomic large deviation bound such as the Hoeffding bound for independent elements~\cite{krmr14}. In our setting, the atomic deviation bound in generic chaining carefully utilizes conditional independence satisfied by the stochastic process $\bxi$. Our analysis for the diagonal terms is based on suitable use of symmetrization, de-symmetrization, and contraction inequalities~\cite{bolm13,leta91}. However, we cannot use the standard form for symmetrization and de-symmetrization which are based on i.i.d.~elements. We generalize the classical symmetrization and de-symmetrization results~\cite{bolm13} to stochastic processes $\bxi$ in our setup, and subsequently utilize these inequalities to bound the diagonal terms.
%
%In particular, we bound $L_p$-norms of two types of random variables respectively based on the contributions from the off-diagonal and diagonal terms in $C_{\cA}(\bxi)$ and develop several new results in the process which may be of independent interest. We consider two types of conditional distributions for $\xi_j|\xi_1,\ldots,\xi_{j-1}$: bounded, and unbounded sub-Gaussian, needing different analysis tools for the diagonal terms.
%
We present a gentle exposition to the analysis in Section~\ref{sec:result} and the technical proofs are all in the Appendix. We have tried to make the exposition self-contained beyond certain key definitions and concepts such as Talagrand's $\gamma$-function and admissible sequence in generic chaining~\cite{tala14}.

%Finally, note that our analysis is for uniform bounds over a set $\cA$ of matrices and is hence considerably more general than Hanson-Wright inequality [], which consider quadratic forms corresponding to a single matrix $A$.

\swcomment{should this be part of section 2 instead?}

{\bf Notation.} Our results are for stochastic processes $\bxi = \{\xi_j\}$ adapted to another stochastic process $F = \{F_i\}$ with both moment and conditional independence assumptions outlined in detail in Section\ref{sec:setup}. We will consider conditional probabilities $X_j = \xi_j | f_{1:j}$, where $f_{1:j}$ is a realization of $F_{1:j}$, and assume $X_j$ to be zero-mean $L$-sub-Gaussian, i.e., $\P(|X_j|>\tau) \leq 2\exp(-\tau^2/L^2)$ for some constant $L > 0$ and all $\tau \geq \tau_0$, a constant~\cite{vers14,vers18}. For the exposition, we will call a random variable sub-Gaussian without explicitly referring to the constant $L$. With $n$ denoting the length of the stochastic process, we will abuse notation and consider a random vector $\bxi = [\xi_j] \in \R^n$ corresponding to the stochastic process $\bxi = \{\xi_j\}$, where the usage will be clear from the context. Our results are based on two classes of complexity measures of a set of $(m \times n)$ matrices $\cA$. The first class, denoted by $d_F(\cA)$ and $d_{2 \rightarrow 2}(\cA)$, are the radius of $\cA$ in the Frobenius norm $\| A \|_F = \sqrt{\tr(A^T A)}$ and the operator norm $\| A \|_{2 \rightarrow 2} = \sup_{\| \x\|_2 \leq 1} \| A \x \|_2$. 
%and the $(2,\infty)$-norm $\| A \|_{2,\infty} = \| \| A_j \|_2 \|_{\infty}$, where $A_j$ is the $j^{th}$ column of $A$. 
For the set $\cA$, we have $d_F(\cA) = \sup_{A \in \cA} \| A \|_F$, and $d_{2 \rightarrow 2}(\cA) = \sup_{A \in \cA} \| A \|_{2 \rightarrow 2}$.
%, and $d_{2,\infty}(\cA) = \sup_{A \in \cA} \| A \|_{2,\infty}$. 
The second class in Talagrand's $\gamma_2(\cA, \| \cdot \|_{2 \rightarrow 2})$ functional, defined in Section~\ref{sec:result}~\cite{tala14,tala05}. Recent literature have used the notion of Gaussian width: $w(\cA) = E \sup_{A \in \cA} | \tr (G^T A)|$ where $G =[g_{i,j}] \in \R^{m \times n}$ have i.i.d.~normal entries, i.e., $g_{i,j} \sim N(0,1)$. It can be shown~\cite{tala14} that $\gamma_2(\cA, \| \cdot \|_{2 \rightarrow 2})$ can be bounded by the Gaussian width $w(\cA)$, i.e., $\gamma_2(\cA, \| \cdot \|_{2 \rightarrow 2}) \leq c w(\cA)$, for some constant $c>0$. Our analysis will be based on bounding $L_p$-norms of suitable random variables. For a random variable $X$, its $L_p$-norm is $\| X \|_{L_p} = (\E|X|^p)^{1/p}$.

%\newpage

\section{Setup and Preliminaries}
\label{sec:setup}

\swedit{In this section we describe the formal set up of stochastic processes for which we provide large deviation bounds.} Let $\bxi  = \{\xi_i\} = \{\xi_1,\ldots,\xi_n\}$ be a sub-Gaussian stochastic process which is decoupled when conditioned on another stochastic process  $F = \{F_i\} = \{F_1,\ldots,F_n\}$. In particular, we assume:
\begin{enumerate}[(SP-1)]
\item for each $i=1,\ldots,n$, $\xi_i |f_{1:i}$ is a zero mean sub-Gaussian random variable~\cite{vers18} for all realizations $f_{1:i}$ of $F_{1:i}$; and
\item for each $i=1,\ldots,n$, there exists an index $\varrho(i) \leq i$ which is non-decreasing, i.e., $\varrho(j) \leq \varrho(i)$ for $j < i$, such that $\xi_i \perp \xi_j | F_{1:\varrho(i)}, j < i$ and $\xi_i \perp F_k | F_{1:\varrho(i)}, k > \varrho(i)$.
\end{enumerate}
where $\perp$ denotes (conditional) independence. The stochastic process $\bxi = \{\xi_i\}$ is said to be {\em adapted to the process $F=\{F_i\}$ satisfying (SP-1) and (SP-2)}. The nomenclature is inspired by the corresponding usage in the context of martingales, we briefly discuss such classical usage and related concepts in Section~\ref{ssec:dtsold}.

(SP-1) is an assumption on the moments of the distributions $\xi_i|f_{1:i}$. Note that the assumption allows the specifics of the distribution to depend on the history. (SP-2) is an assumption on the conditional independence structure of $\bxi$. The assumption allows $\xi_i$ to depend on the history $F_{1:\varrho(i)}$. Further, we can have $F_{i-1}$ depend on $\xi_{i-1}$ and $\xi_{i}$ depend on $F_{i-1}$. In the sequel, we give concrete examples of graphical models which follow (SP-1) and (SP-2) and allow different types of dependencies among the random variables. We also give concrete examples of potential interest in the context of machine learning in Section~\ref{sec:app}.

\begin{figure}[t]
\centering
 \includegraphics[width = 0.6 \textwidth]{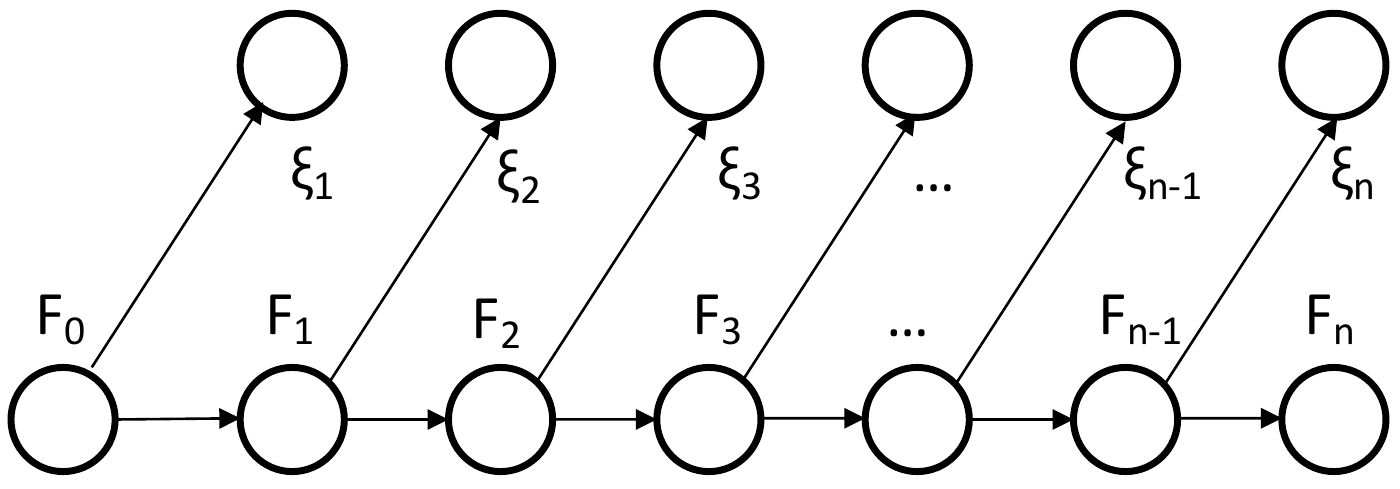}
 \caption[]{Graphical Model 1 (GM1) structure for stochastic process $\{\xi_i\}$ adapted to $\{ F_i\}$ satisfies (SP-2) by construction (Proposition~\ref{prop:gm1}). While we show arrows only from one random variable, e.g., $F_{i-1} \rightarrow \xi_i$, the conditional random variable $\xi_i|F_{1:(i-1)}$ can have dependence on the entire history $F_{1:(i-1)}$. All these arrows are not depicted in this and other figures to avoid clutter.}
\label{fig:ci}
\end{figure}

\begin{figure}[t]
\centering
 \includegraphics[width = 0.6 \textwidth]{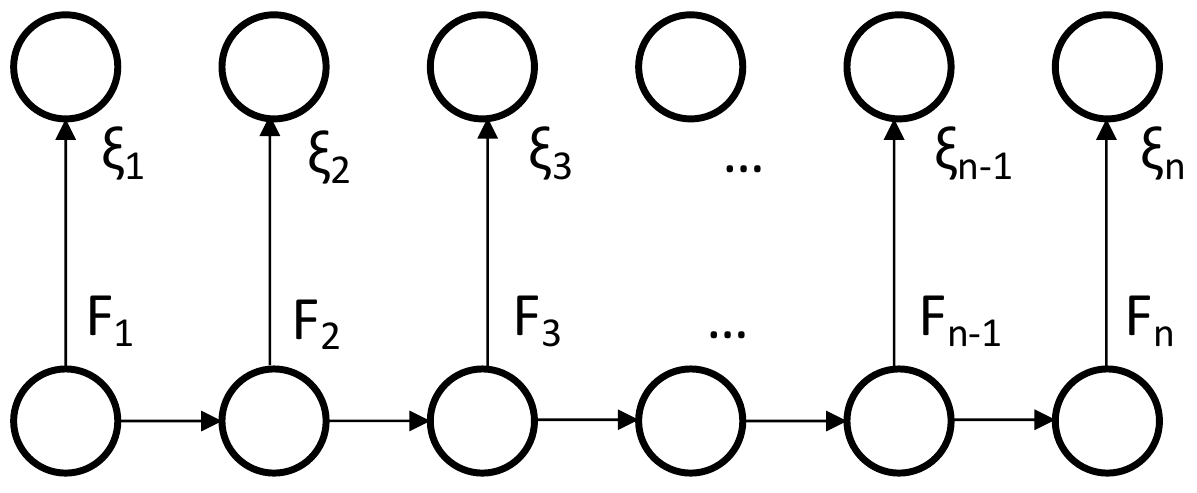}
 \caption[]{
 Graphical Model 2 (GM2) structure for stochastic process $\{\xi_i\}$ adapted to $\{ F_i\}$ satisfies (SP-2) by construction (Proposition~\ref{prop:gm2}). While we show arrows only from one random variable, e.g., $F_i \rightarrow \xi_i$, the conditional random variable $\xi_i|F_{1:i}$ can have dependence on the entire history $F_{1:i}$.}
\vspace*{-4mm}
\label{fig:hmm}
\end{figure}

\begin{figure}[t]
\centering
 \includegraphics[width = 0.6 \textwidth]{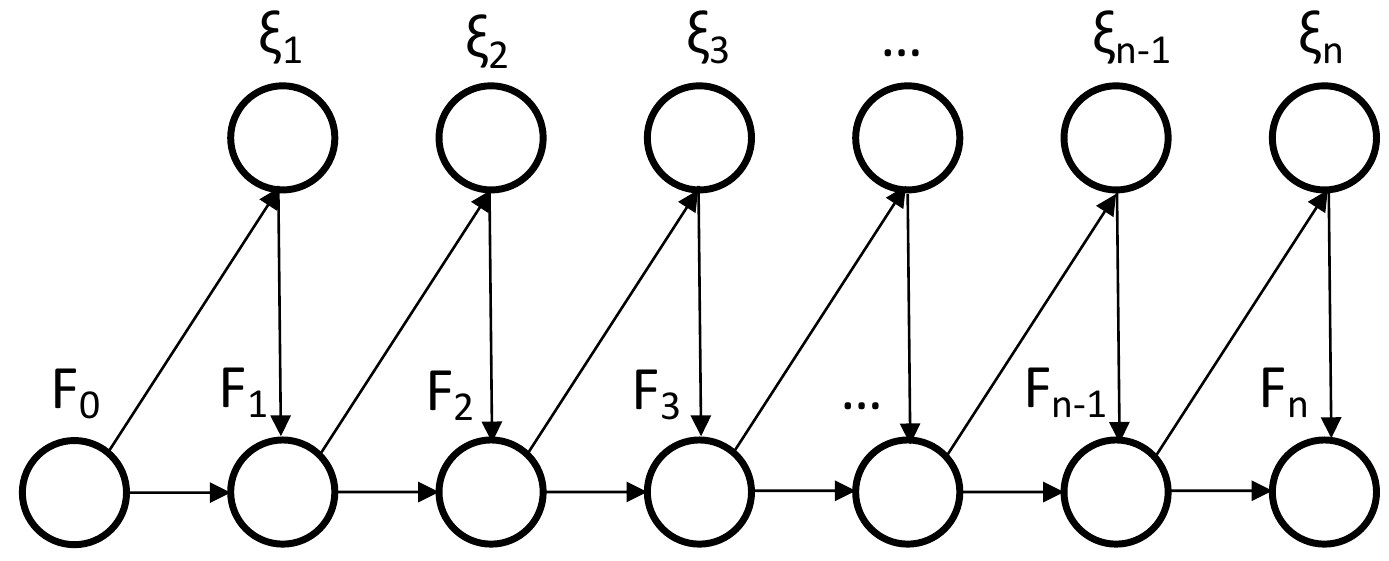}
 \caption[]{
 Graphical Model 3 (GM3) structure for stochastic process $\{\xi_i\}$ adapted to $\{ F_i\}$ satisfies (SP-2) by construction (Proposition~\ref{prop:gm3}). Note that there is no restriction on the conditional distribution $F_i \mid (F_{1:(i-1)},\xi_i)$, so that $F_i$ can have arbitrary dependence on $F_{1:(i-1)}$ and $Z_i$.
While we show arrows only to one random variable, e.g., $F_{i-1} \rightarrow \xi_i$, the conditional random variable $\xi_i|F_{1:(i-1)}$ can have dependence on the entire history $F_{1:(i-1)}$. Similarly, $F_i|F_{1:(i-1)},Z_i$ is illustrated only with arrows from $F_{i-1},Z_i$ to $F_i$ to avoid clutter. }
\vspace*{-4mm}
\label{fig:cilcb}
\end{figure}

Examples of graphical models satisfying (SP-2) are shown as Graphical Model 1 (GM1) in Figure~\ref{fig:ci}, Graphical Model 2 (GM2) in Figure~\ref{fig:hmm}, and Graphical Model 3 (GM3) in Figure~\ref{fig:cilcb}. For GM1, $\varrho(i)=i-1$ and $F_i$ depends on $F_{1:(i-1)}$, but not on $\xi_i$. Further, $\xi_i$ can depend on the entire history $F_{1:(i-1)}$. GM2 is a variant of GM1 and structurally looks like a HMM (hidden Markov model) with $\varrho(i)=i$, $F_i$ depending on $F_{i-1}$ (or the entire history $F_{1:(i-1)}$), and $\xi_i$ depends on $F_i$ (or the entire history $F_{1:i}$). GM3 is a more complex model with $\varrho(i)=i$ and $F_i$ depends both on $F_{1:(i-1)}$ and $\xi_i$. For GM1 and GM3, we consider an additional `prior' $F_0$, and the properties (SP-1) and (SP-2) can be naturally extended to include such a prior. An interesting special case of interest for GM3 is when $\xi_i|F_{1:(i-1)}$ is centered sub-Gaussian and $F_i$ is a deterministic function of $(F_{i-1},\xi_i)$, i.e., $F_i = \zeta(F_{i-1},\xi_i)$. Note that the distribution  
\begin{equation*}
    \P(\xi_i | F_{1:i}) = \begin{cases}
        \P(\xi_i | F_{1:(i-1)})~, & ~~~\text{if}~~~ F_i = \zeta(F_{i-1},\xi_i)~,\\
        0~, & ~~~\text{otherwise}~.
    \end{cases}
\end{equation*}
In other words, a realization $f_{1:n}$ following $f_i = \zeta(f_{1:(i-1)},\xi_i)$ will have $\P(\xi_i|f_{1:i}) = \P(\xi_i|f_{1:(i-1)})$ and will therefore be centered sub-Gaussian if $\P(\xi_i|f_{1:(i-1)})$ is centered sub-Gaussian, which is easy to ensure by design. For certain graphical models, it may be at times more natural to first construct a stochastic process $\{Z_i\}$ respecting the graphical model structure governed by (SP-2), and then construct the sequence $\{\xi_i\}$ by conditional centering, i.e., $\xi_i|F_{1:i} = Z_i|F_{1:i} - \E[Z_i|F_{1:i}]$ so that $\E[\xi_i|F_{1:i}] = 0$ as required by (SP-1). Such a centered construction is inspired by how one can construct martingale difference sequences (MDSs) from martingales~\cite{will91}. 

%The specific case where $F_i$ is a deterministic function of $(F_{i-1},\xi_i)$ includes important stochastic processes such as martingale difference sequences (MDSs) [][] and recent work on smoothed analysis of contextual linear bandits [][]. 

Next we show that for the graphical models GM1, GM2, and GM3, the conditional independence assumption (SP-2) above is satisfied by construction based on the graph structure. We start by recalling the definitions of $d$-separation and $d$-connection~\cite{pear88,barb12}.
\begin{defn}[$d$-connection, $d$-separation] 
Let $X, Y, Z$ be disjoint sets of vertices in a directed graph $G$. $X, Y$ is {\em d-connected} by $Z$ if and only if there exists an undirected path $U$ between some $x \in X, y \in Y$ such that (a) for every collider $C$ on $U$, either $C$ or a descendent of $C$ is in $Z$, and
(b) no non-collider on $U$ is in $Z$. Otherwise $X$ and $Y$ are {\em d-separated} by $Z$.
\end{defn} 
We also recall that $d$-separation implies conditional independence~\cite{barb12,pear88}.
\begin{theo}
If $Z$ d-separates $X$ and $Y$, then $X \perp Y | Z$ for all distributions represented by the graph.
\end{theo}
We will use $d$-separation to show that GM1, GM2, and GM3 satisfy the assumption (SP-2).
\begin{prop}
The graphical models GM1 in Figure~\ref{fig:ci} satisfies (SP-2).
\label{prop:gm1}
\end{prop}
\proof For GM1, we have $\varrho(i)=i-1$. Since GM1 is a tree structured model with no loops, for any $j < i$, there is only one undirected path connecting $\xi_j$ and $\xi_i$, all nodes $F_{j-1},\ldots,F_{i-1}$ in that path are non-colliders, and we are conditioning on them. Thus $\xi_j$ and $\xi_i$ are $d$-separated by $F_{(j-1):(i-1)}$ and hence $\xi_i \perp \xi_j | F_{1:(i-1)}$. Further, for any $k > (i-1)$, there is only one undirected path connecting $\xi_i$ and $F_k$, all nodes $F_{i-1},\ldots,F_{k-1}$ in that path are non-colliders. Thus, if we are conditioning on $F_{i-1}$, $\xi_i$ and $F_k$ are $d$-separated by $F_{(i-1)}$ and hence $\xi_i \perp F_k | F_{1:(i-1)}$. That completes the proof. \qed

\begin{prop}
The graphical models GM2 in Figure~\ref{fig:hmm} satisfies (SP-2).
\label{prop:gm2}
\end{prop}
\proof For GM2, we have $\varrho(i)=i$. Since GM2 is a tree structured model with no loops, for any $j < i$, there is only one undirected path connecting $\xi_j$ and $\xi_i$, all nodes $F_{j},\ldots,F_{i}$ in that path are non-colliders, and we are conditioning on them. Thus $\xi_j$ and $\xi_i$ are $d$-separated by $F_{(j):(i)}$ and hence $\xi_i \perp \xi_j | F_{1:i}$. Further, for any $k > i$, there is only one undirected path connecting $\xi_i$ and $F_k$, all nodes $F_{i},\ldots,F_{k-1}$ in that path are non-colliders. Thus, if we are conditioning on $F_{i}$, $\xi_i$ and $F_k$ are $d$-separated by $F_{i}$ and hence $\xi_i \perp F_k | F_{1:i}$. That completes the proof. \qed

\begin{prop}
The graphical models GM3 in Figure~\ref{fig:cilcb} satisfies (SP-2).
\label{prop:gm3}
\end{prop}
\proof For GM3, we have $\varrho(i)=i$. For any $j < i$, there are $4$ undirected paths connecting $\xi_j$ and $\xi_i$ and we consider each one of them. For the paths $\xi_j,F_{j-1},F_j,\ldots,F_{i-1},\xi_i$ and $\xi_j,F_j,\ldots,F_{i-1},\xi_i$, the intermediate nodes are all non-colliders and we are conditioning on them. For the paths $\xi_j,F_{j-1},F_j,\ldots,F_{i-1},F_i,\xi_i$ and $\xi_j,F_j,\ldots,F_{i-1},F_i,\xi_i$, $F_i$ is a collider but there is at least one non-collider (e.g., $F_j$) and we conditioning on both the collider and the non-collider(s). Thus, $\xi_j$ and $\xi_i$ are $d$-separated given the intermediate nodes, implying $\xi_i \perp \xi_j | F_{1:i}$ since $F_{1:(j-2)}$ is not part of any path. Further, for any $k > i$, there are two undirected paths connecting $\xi_i$ and $F_k$: $\xi_i,F_{i-1},F_i,\ldots,F_k$ and $\xi_i,F_i,\ldots,F_k$.
All intermediate nodes in each path are non-colliders, and we are conditioning on one \qgedit{of} them: $F_i$. Thus $\xi_j$ and $\xi_i$ are $d$-separated by $F_{(j):(i)}$ and hence $\xi_i \perp \xi_j | F_{1:i}$.
That completes the proof. \qed

Next, we show that if a model satisfies (SP-2), then $\{\xi_i\}$ is conditionally independent given $\{ F_i\}$. 
\begin{prop}
For a graphical model which satisfies (SP-2), we have
\begin{equation}
    \P( \xi_{1:n} \mid F_{1:n}) = \prod_{i=1}^n \P(\xi_i \mid F_{1:\varrho(i)}) = \prod_{i=1}^n \P(\xi_i \mid F_{1:n}) ~.
\end{equation}
\label{prop:sp2ci}
\end{prop}
\proof We prove the statement recursively, by starting from $\xi_n$ and stepping backwards. For $i=n$ with corresponding $\varrho(i)=\varrho(n)$, since $\xi_n \perp \xi_{1:(n-1)} | F_{1:\varrho(n)}$ by (SP-2), we have
\begin{align*}
    \P(\xi_{1:n} | F_{1:n}) & =  \P(\xi_{1:(n-1)}, \xi_n | F_{1:\varrho(n)}, F_{\varrho(n)+1:n}) \\
    & = \P(\xi_{1:(n-1)} | F_{1:\varrho(n)}, F_{\varrho(n)+1:n} ) \P(\xi_n | F_{1:\varrho(n)}, F_{\varrho(n)+1:n} ) \\
    & =  \P(\xi_{1:(n-1)} | F_{1:n}) \P(\xi_n | F_{1:\varrho(n)})~,
\end{align*}
since $\xi_n \perp F_{\varrho(n)+1:n} | F_{1:\varrho(n)}$ by (SP-2).
Repeating the same argument for $i=(n-1),\ldots,1$ completes the proof for the first part. The second part follows since $\xi_i \perp F_{\varrho(i)+1:n} | F_{1:\varrho(i)}$ by (SP-2) so that $\P(\xi_i|F_{1:\varrho(i)}) = \P(\xi_i|F_{1:n})$. \qed 

\begin{figure}[t]
\centering
 \includegraphics[width = 0.6 \textwidth]{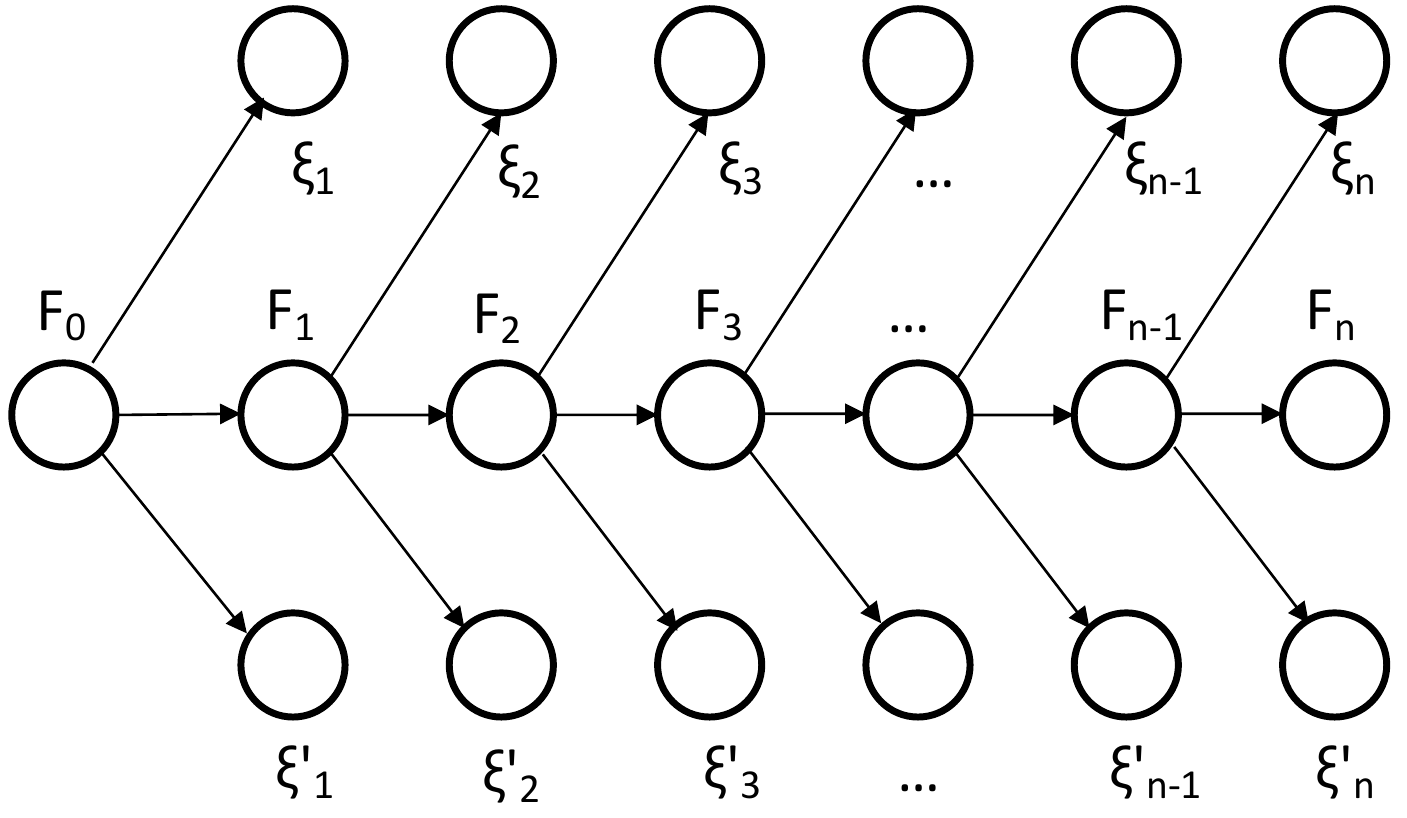}
 \caption[]{Graphical Model 1 (GM1) from Figure~\ref{fig:ci} with the decoupled tangent sequence (DTS) $\{\xi'_i \}$ following Definition~\ref{defn:dts}. For GM1, since the index $\varrho(i) = i-1$ from (SP-2), the DTS $\{\xi'_i\}$ satisfies $\P(\xi_i | F_{1:i}) = \P(\xi'_i | F_{1:i})$ and also $\P(\xi_i | F_{1:i}) = \P(\xi_i | F_{1:(i-1)})$ and $\P(\xi'_i | F_{1:i}) = \P(\xi'_i | F_{1:(i-1)})$ by (SP-2).
 Further,  $\xi_i \perp \xi'_i | F_{1:i}$ and also  $\xi_i \perp \xi'_i | F_{1:(i-1)}$ by (SP-2).
As in Figure~\ref{fig:ci}, while we show arrows only to one r.v., e.g., $F_{i-1} \rightarrow \xi'_i$, the conditional random variable $\xi'_i|F_{1:i-1}$ can have dependence on the entire history $F_{1:i-1}$. }
\label{fig:tgtci}
\end{figure}

\begin{figure}[t]
\centering
 \includegraphics[width = 0.6 \textwidth]{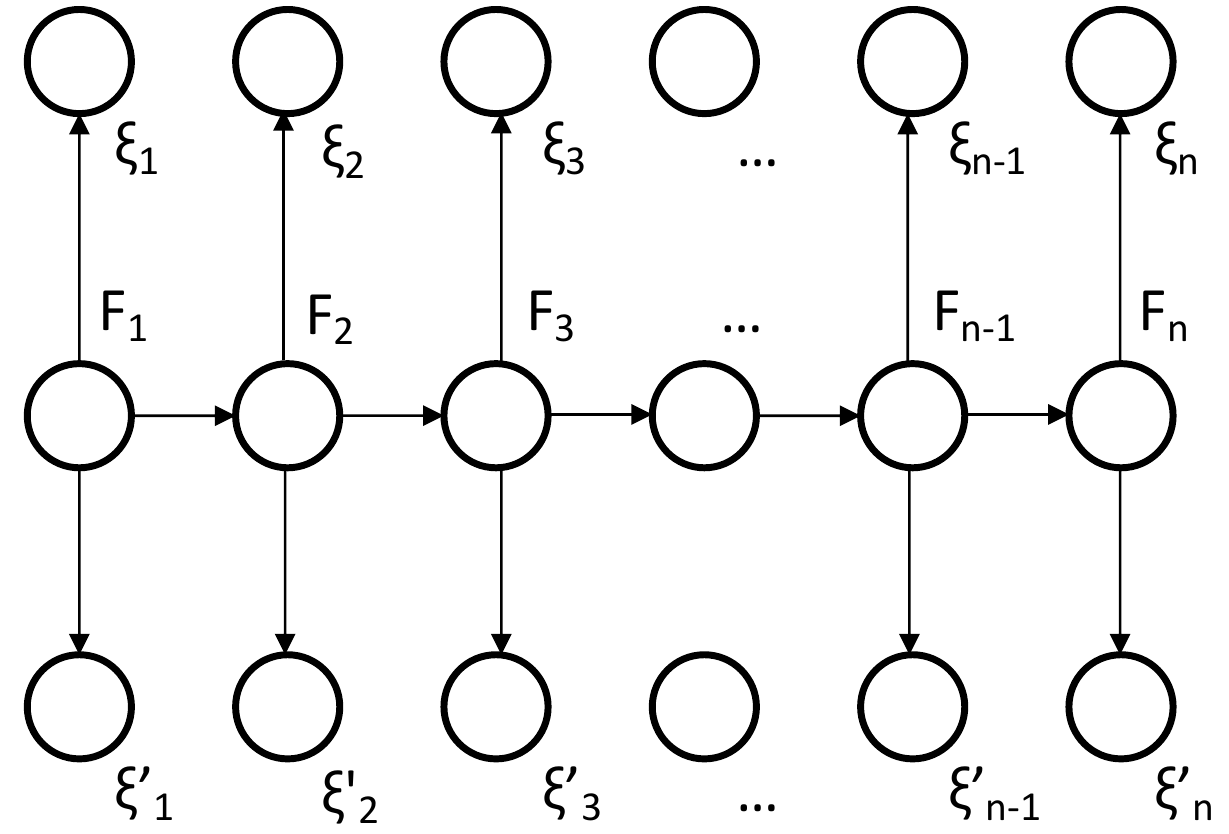}
 \caption[]{Graphical Model 2 (GM2) from Figure~\ref{fig:hmm} with the decoupled tangent sequence (DTS) $\{\xi'_i \}$ following Definition~\ref{defn:dts}. For GM2, since the index $\varrho(i) = i$ from (SP-2), the DTS $\{\xi'_i\}$ satisfies $\P(\xi_i | F_{1:i}) = \P(\xi'_i | F_{1:i})$ and $\xi_i \perp \xi'_i | F_{1:i}$.  As in Figure~\ref{fig:hmm}, while we show arrows only to one r.v., e.g., $F_{i} \rightarrow \xi'_i$, the conditional random variable $\xi'_i|F_{1:i}$ can have dependence on the entire history $F_{1:i}$.}
 \vspace*{-4mm}
\label{fig:tgthmm}
\end{figure}

\begin{figure}[t]
\centering
 \includegraphics[width = 0.6 \textwidth]{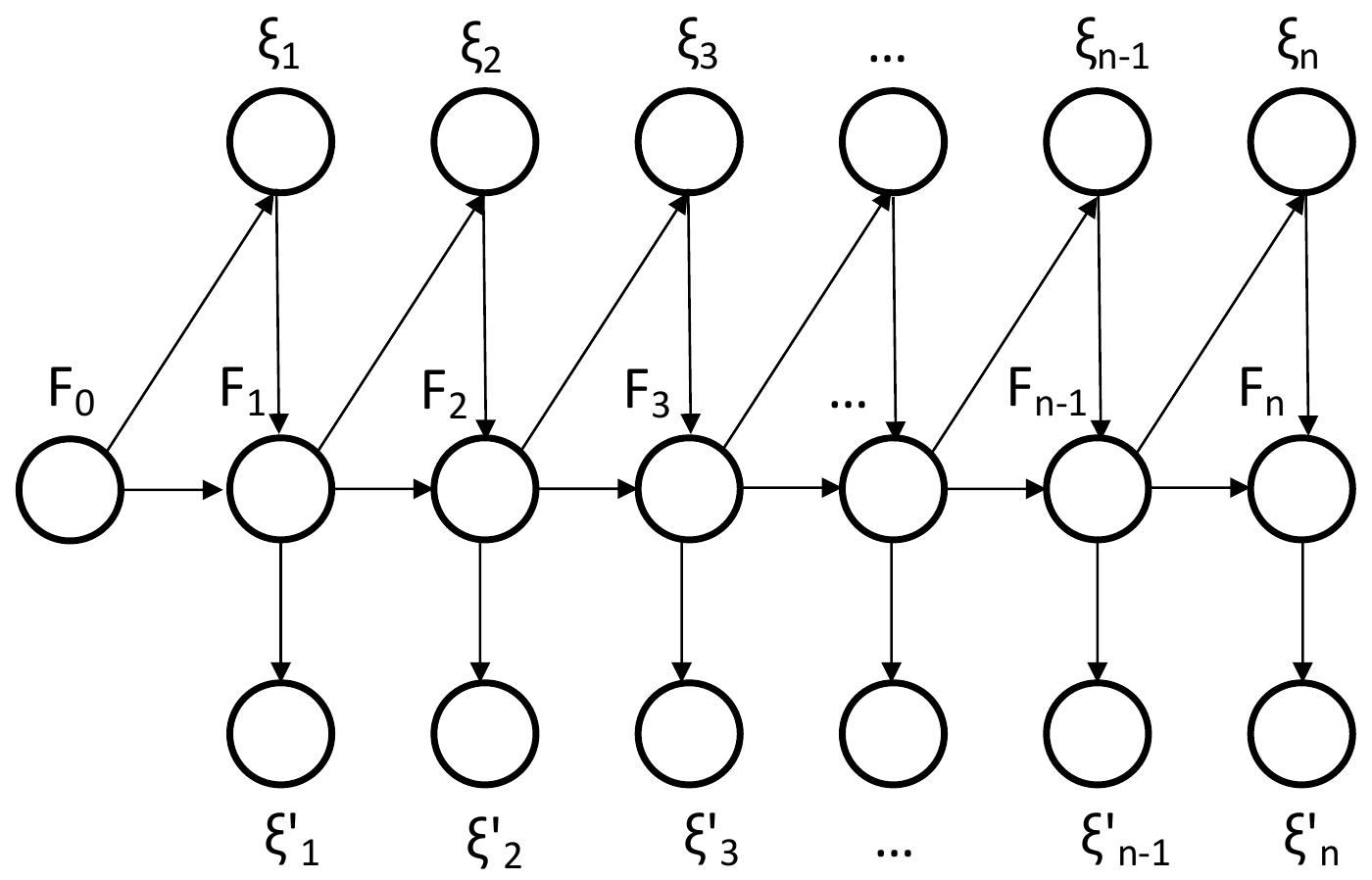}
 \caption[]{Graphical Model 3 (GM3) from Figure~\ref{fig:cilcb} with the decoupled tangent sequence (DTS) $\{\xi'_i \}$ following Definition~\ref{defn:dts}. For GM3, since the index $\varrho(i) = i$ from (SP-2), the DTS $\{\xi'_i\}$ satisfies $\P(\xi_i | F_{1:i}) = \P(\xi'_i | F_{1:i})$ and $\xi_i \perp \xi'_i | F_{1:i}$.  Note that for GM3, $\P(\xi_i | F_{1:i}) \propto \P(\xi_i | F_{1:(i-1)}) \P(F_i|\xi_i,F_{1:(i-1)}$ is the posterior distribution, and we construct the DTS $\{\xi'_i\}$ so that $\P(\xi'_i | F_{1:i}) = \P(\xi_i | F_{1:i})$. As in Figure~\ref{fig:cilcb}, while we show arrows only to one r.v., e.g., $F_{i} \rightarrow \xi'_i$, the conditional random variable $\xi'_i|F_{1:i}$ can have dependence on the entire history $F_{1:i}$.}
\vspace*{-4mm}
\label{fig:tgtcilcb}
\end{figure}

For any stochastic process $\bxi = \{\xi_i\}$ adapted to $F = \{F_i\}$ satisfying (SP-1) and (SP-2), for our analysis we will consider another stochastic process $\bxi' = \{\xi'_i\}$ called a {\em decoupled tangent sequence} (DTS). The name is inspired by a closely related idea in the classical literature on decoupling~\cite{pegi99}, and we present a brief exposition to this classical usage in Section~\ref{ssec:dtsold}.
\begin{defn}[Decoupled Tangent Sequence (DTS)]
For any stochastic process $\{\xi_i\}$ satisfying (SP-1) and (SP-2) based on another process $\{F_i\}$, we define a stochastic process $\{ \xi'_i\}$ to be a decoupled tangent sequence (DTS) if
\begin{enumerate}[(DTS-1)]
    \item $\xi_i \perp \xi'_i | F_{1:i}$ for $i=1,\ldots,n$; and
    \item $\P(\xi_i | F_{1:i}) = \P(\xi'_i | F_{1:i})$ for $i=1,\ldots,n$.
\end{enumerate}
\label{defn:dts}
\end{defn}
In other words, the process $\{\xi'_i\}$ is componentwise conditionally independent and conditionally identically distributed with respect to $\{\xi_i\}$ where the conditioning is over $F_{1:i}$ for each $i=1,\ldots,n$. Figures~\ref{fig:tgtci},  \ref{fig:tgthmm}, and \ref{fig:tgtcilcb} show the extended graphical models for GM1, GM2, and GM3 which include the DTS $\{\xi'_i\}$. Note that such a DTS $\{ \xi_i\}$ can be constructed for any process $\{\xi_i\}$ satisfying (SP-1) and (SP-2) by simply making each $\xi'_i$ conditionally i.i.d.~w.r.t~$\xi_i$ conditioned on $F_{1:i}$. The figures show examples of such constructions. Further, by construction, $\{\xi'_i\}$ is a stochastic process adapted to $\{ F_i\}$ and satisfies (SP-1) and (SP-2) with $\varrho(i) = i$.

\section{Main Results}
\label{sec:result}

Let $\cA$ be a set of $(m \times n)$ matrices and let $\bxi$ be a stochastic process adapted to $F$ satisfying (SP-1) and (SP-2). The random variable of interest for the current analysis is:
\begin{equation}
    C_{\cA}(\bxi)  \triangleq ~\sup_{A \in \cA} \left| \|A\bxi\|_2^2 - \E\|A\bxi\|_2^2 \right|~. 
    \label{eq:caxi}
\end{equation}
Based on the literature on empirical processes and generic chaining \cite{tala14,ledo13}, the random variable $C_{\cA}(\bxi)$ can be referred to as an order-2 sub-Gaussian chaos~\cite{tala14,krmr14}. Unlike the Hanson-Wright inequality \cite{hawr71,ruve13}, which also considers large deviation bounds for quadratic forms of random vectors with a fixed matrix $A$, our focus is on an uniform bound over an entire set of matrices $\cA$ and $\bxi$ is a stochastic process as opposed to vectors with independent entries in the current literature~\cite{vers18}. 

While widely used results like the restricted isometry property (RIP) \cite{cata05,dono06} and Johnson-Lindenstrauss (J-L) lemma \cite{joli84,wood14} do not explicitly appear in the above form, getting such results from a large deviation bound on $C_{\cA}(\bxi)$ is straightforward~\cite{krmr14,merw18}. In particular, to get results like RIP and J-L, we need to make a conversion: the matrix $A$ typically gets vectorized to $\text{vec}(A)$ in a suitable (restricted) set $\cA$ and the random vector $\bxi$ gets converted into a suitable random matrix $X$ (see Section~\ref{sec:app} for details). For ease of exposition, we will refer to such converted but otherwise equivalent form as the random matrix form of $C_{\cA}(\bxi)$.

State-of-the-art results on large deviation bounds on $C_{\cA}(\bxi)$ only consider $\bxi$ with {\em independent} sub-Gaussian entries~\cite{krmr14}. Such independence in the order-2 chaos form gets converted to row-wise or entry-wise independence in the random matrix form, e.g., for RIP type results \cite{krmr14,ledo13,cata05}. 

%\abcomment{needs major edits from this point onwards, ongoing} 

\subsection{The Main Result: Warm-up}
The main technical result in the paper is a large deviation bound on $C_{\cA}(\bxi)$ for the setting when $\bxi$ is a 
stochastic process adapted to $F$ satisfying (SP-1) and (SP-2), as defined in Section~\ref{sec:setup}. As illustrated through the example graphical models GM1, GM2, and GM3, $\bxi$ has statistically dependent rows and the rows can even be adaptively generated as illustrated by GM3. 

% sub-Gaussian martingale difference sequence (MDS) \cite{will91,vers18}. In particular, if $\Xi = \{ \xi_j\}$ is the sequence of random variables adapted to an increasing sequence of $\sigma$-fields $\{ \cF_j\}$, then
% \begin{enumerate}[(MDS-1)]
%     \item $\xi_j|\cF_{j-1}$ is a sub-Gaussian random variable~\cite{vers14}; and
%     \item $E[\xi_j|\cF_{j-1}] = 0$.
% \end{enumerate}
% We emphasize that $\bXi$ being a MDS allows for arbitrary dependence of the zero-mean distribution $p(\xi_j|\cF_{j-1})$ on the history. Consider the factorization of a general joint distribution:
% \begin{align*}
%     p(\xi_1,\xi_2,\ldots,\xi_j,\ldots,\xi_n) & = p(\xi_1) p(\xi_2|\xi_1) \cdots p(\xi_j|\xi_1,\ldots,\xi_{j-1}) \cdots p(\xi_n|\xi_1,\ldots,\xi_{n-1})~.
% \end{align*}
% The only restriction the sub-Gaussian MDS assumption puts on such general joint distribution is that the conditional distributions $\xi_j|\xi_1,\ldots,\xi_{j-1}$ be zero-mean sub-Gaussian~\cite{vers14}. As long as this simple restriction is satisfied, the dependencies can in fact be arbitrary. In particular, there is no need for statistical independence among the entries. Such a perspective illustrates the generality of our setting and results.

To develop large deviation bounds on $C_{\cA}(\bxi)$, we decompose the quadratic form into terms depending on the off-diagonal and the diagonal elements of $A^T A$ respectively as follows:
\begin{align}
    B_{\cA}(\bxi) & \triangleq ~\sup_{A \in \cA} \left| \sum_{\substack{j,k=1\\j \neq k}}^n \xi_j \xi_k \langle A_j, A_k \rangle \right| ~, \label{eq:baxi} \\
    D_{\cA}(\bxi) & \triangleq ~\sup_{A \in \cA} \left| \sum_{j=1}^n (|\xi_j|^2 - \E|\xi_j|^2) \| A_j \|_2^2 \right| ~. \label{eq:daxi}
\end{align}
Note that the contributions from the off-diagonal terms of $A^T A$ to $\E\|A\bxi\|_2^2$ is 0. To see this, with  $A_j$ denoting the $j^{th}$ column of $A$, by linearity of expectation we have
\begin{align*}
    \E_{\bxi} \left[ \sum_{\substack{j,k=1\\j\neq k}}^n \xi_j \xi_k \langle A_j, A_k \rangle \right]
    & =  \sum_{\substack{j,k=1\\j\neq k}}^n  \E_{\xi_j,\xi_k} [\xi_j \xi_k] \langle A_j, A_k \rangle 
     = \sum_{\substack{j,k=1\\j\neq k}}^n \E_{F_{1:n}}\left[ \E_{\xi_j,\xi_k|F_{1:n}} \left[\xi_j \xi_k \right]\right] \langle A_j, A_k \rangle  \\
& \stackrel{(a)}{=} \sum_{\substack{j,k=1\\j\neq k}}^n \E_{F_{1:n}}\left[ \E_{\xi_j|F_{1:n}}[\xi_j] \E_{\xi_k|F_{1:n}}[\xi_k] \right]  \langle A_j, A_k \rangle  \stackrel{(b)}{=} 0~,
\end{align*}
where (a) follows since $\xi_j \perp \xi_k | F_{1:n}$ by (SP-2) and Proposition~\ref{prop:sp2ci}, and (b) follows since $\E_{\xi_j|F_{1:n}}[\xi_j] = \E_{\xi_k|F_{1:n}}[\xi_k] =0$ by (SP-1). As a result, the contributions from the off-diagonal elements of $A^T A$ in $\E\|A\bxi\|_2^2$ is zero.

Now, by definition and Jensen's inequality, we have 
\begin{align*}
    C_{\cA}(\bxi) & = ~ \sup_{A \in \cA} \left| \|A\bxi\|_2^2 - \E\|A\bxi\|_2^2 \right| \\
    & =~ \sup_{A \in \cA} \left| \sum_{\substack{j,k=1\\j \neq k}}^n \bxi_j \bxi_k \langle A_j, A_k \rangle + \sum_{j=1}^n (|\xi_j|^2 - \E|\xi_j|^2) \| A_j \|_2^2 \right| \\
    & \leq~ \sup_{A \in \cA} \left| \sum_{\substack{j,k=1\\j \neq k}}^n \bxi_j \bxi_k \langle A_j, A_k \rangle \right| + \sup_{A \in \cA} \left| \sum_{j=1}^n (|\xi_j|^2 - \E|\xi_j|^2) \| A_j \|_2^2 \right| \\
    & =~ B_{\cA}(\bxi) +  D_{\cA}(\bxi) 
\end{align*}    
Therefore, for any $p \in [1,\infty)$, we have 
\begin{align}
    \| C_{\cA}(\bxi) \|_{L_p} & \leq \| B_{\cA}(\bxi)\|_{L_p}  +  \| D_{\cA}(\bxi) \|_{L_p}~.
\end{align}
Our approach to getting a large deviation bound for $C_{\cA}(\bxi)$ is based on bounding
$\| C_{\cA}(\bxi) \|_{L_p}$, which in turn is based on bounding $\| B_{\cA}(\bxi) \|_{L_p}$ and $\| D_{\cA}(\bxi) \|_{L_p}$. Such bounds lead to a bound on $\| C_{\cA}(\bxi)\|_{L_p}$  of the form
\begin{equation}
    \| C_{\cA}(\bxi)\|_{L_p} \leq a + \sqrt{p} \cdot b + p \cdot c~, \quad \forall p \geq 1~,
\end{equation}
where $a,b,c$ are constants which do not depend on $p$.
Note that by using the moment-generating function and Markov's inequality \cite{will91,vers12}, these $L_p$-norm bounds imply, for all $u >0$
\begin{equation}
    P(\left| C_{\cA}(\bxi) \right|  \geq a + b \cdot \sqrt{u} + c \cdot u ) \leq e^{-u}~,
    \label{eq:mark1}
\end{equation}
or, equivalently 
\begin{equation}
    P(\left| C_{\cA}(\bxi) \right|  \geq a + u ) \leq \exp\left\{-\min\left(\frac{u^2}{4b^2}, \frac{u}{2c}\right)\right\}~,
    \label{eq:mark2}
\end{equation}
which yields the desired large deviation bound. 
%\sw{is there a citation for above? or just say this follows from moment generating fn?} \ab{Yes, this follows from the moment generating  function -- will add cites}

The analysis for bounding the $L_p$ norms of $C_{\cA}(\bxi)$ for any $p \geq 1$ will thus be based on bounding the $L_p$ norms of $B_{\cA}(\bxi)$, a term based on the off-diagonal elements of $A^T A$, and that of $D_{\cA}(\bxi)$, a term based on the diagonal elements of $A^T A$. For convenience, we will refer to $B_{\cA}(\bxi)$ as the off-diagonal term and $D_{\cA}(\bxi)$ as the diagonal term. We now discuss how we will construct the bounds on $\| B_{\cA}(\bxi)\|_{L_p}$ and $\| D_{\cA}(\bxi) \|_{L_p}$.

\subsection{The Off-diagonal Term $B_{\cA}($\texorpdfstring{$\boldsymbol{\xi}$}{\textxi}$)$} 
For the off-diagonal term $B_{\cA}(\bxi)$, the bound on $\| B_{\cA}(\bxi)\|_{L_p}$ is based on two techniques: decoupling~\cite{pegi99} and generic chaining~\cite{tala14}. In our context, since $\bxi$ is a stochastic process, we need to extend certain key results in both of these themes to be applicable to $\bxi$ satisfying (SP-1) and (SP-2). Our main result in decoupling, stated below, extends the classical result for $\bxi$ with i.i.d.~entries to stochastic processes $\bxi$ satisfying (SP-1) and (SP-2). We prove the result in Appendix~\ref{sec:mart-dec}.
\begin{theo}
Let $\bxi = \{\xi_i\}$ be a stochastic process adapted to $F=\{ F_i \}$ satisfying (SP-1) and (SP-2). Let $\bxi' = \{ \xi'_i \}$ be any decoupled tangent sequence to $\bxi = \{ \xi_i \}$ so that (DTS-1) and (DTS-2) are satisfied.
%$\P(\xi_i | F_{1:(i-1)}) = \P(\xi'_i | F_{1:(i-1)})$ and $\xi_i \perp \xi'_i \mid F_{1:(i-1)}$ (Figure~\ref{fig:tgtci}).
%, with $\{ \xi'\}$  being conditionally independent given $\cG = \sigma(\{d_i\})$.
Let ${\cal B}$ be a collection of $(n \times n)$ symmetric matrices. Let $h: \R \mapsto \R$ be a convex function. Then,
\beq
\E_{\bxi} \left[ \sup_{B \in {\cal B}} h \left( \sum_{\substack{j,k=1\\ j\neq k}}^n \xi_j \xi_k B_{j,k} \right) \right]
\leq 4 \E_{\bxi,\bxi'}\left[ \sup_{B \in {\cal B}} h \left( \sum_{\substack{j,k=1}}^n \xi_j \xi'_k B_{j,k} \right) \right]~.
\label{eq:mds-dec0}
\eeq
\label{th:mds-dec0}
\end{theo}
The key benefit from decoupling is that rather than working with a quadratic form of $\bxi$ without contributions from the diagonal elements, we will be working with the decoupled conditionally linear forms on $\bxi$ and $\bxi'$ where we will be able to use more standard results like Hoeffding bounds~\cite{bolm13} under suitable conditioning. For our analysis, the convex function $h(\cdot)$ in Theorem~\ref{th:mds-dec0} will be $L_p$ norms for $p \geq 1$.

The second part of the analysis for bounding $\| B_{\cA}(\bxi)\|_{L_p}$ uses generic chaining~\cite{tala14}. The focus of the analysis will be to bound the right hand side of \eqref{eq:mds-dec0} in Theorem~\ref{th:mds-dec0} with $B_{j,k} = \langle A_j, A_k\rangle$. First note that a naive approach to 
doing such a bound would end up involving the cardinality of $\cB$ in Theorem~\ref{th:mds-dec0} or equivalently cardinality of $\cA$ for our analysis because of the $\sup_{A \in \cA}$. Such bounds will be useless for most interesting sets $\cA$, e.g., set of sparse or low-rank matrices. Generic chaining can fully exploit any structure in $\cA$ based on a hierachical decomposition~\cite{tala14,tala05} and is arguably one of the most powerful tools for such analysis. Since we use generic chaining, the results are in terms of Talagrand's $\gamma$-functions. We need the following key definition due to Talagrand~\cite{tala14}. 

%\swcomment{is it possible to briefly explain what generic chaining is used for? this was highlighted as a key feature}\ab{to do}
\begin{defn}
For a metric space $(T,d)$, an admissible sequence of $T$ is a collection of subsets of $T$, $\{T_r : r \geq 0\}$,
with $|T_0| = 1$ and $|T_r| \leq 2^{2^r}$ for all $r \geq 1$. For $\beta \geq 1$, the $\gamma_{\beta}$ functional is defined by
\beq
\gamma_{\beta}(T,d) = \inf \sup_{t \in T} \sum_{r=0}^{\infty} 2^{r/\beta} d(t,T_r)~,
\eeq
where the infimum is over all admissible sequences of $T$.
\end{defn}
In particular, our results are in terms of $\gamma_2(\cA, \| \cdot \|_{2 \rightarrow 2})$, which is related to the Gaussian width of the set by the majorizing measure theorem~\cite[Theorem 2.1.1]{tala05}\cite[Theorem 2.4.1]{tala14}. 
%\ab{the result really uses a $L_2$ metric, which will be $\| \cdot \|_F$ (Frobenius norm), whereas our results use $\| \|_{2 \rightarrow 2}$ (operator norm), which is tighter.}
Recent years have seen major advances in using Gaussian width for both statistical and computational analysis in the context of high-dimensional statistics and related areas \cite{crpw12,bcfs14,oyrs18,chba15}. Hence, recent tools for bounding Gaussian width \cite{crpw12,chba15} can be applied to our setting to get concrete bounds for cases of interest. For example, \abcomment{please check} if $\cA$ is a set of $s$-sparse $(m \times n)$ matrices, $\gamma_2(\cA, \| \cdot \|_{2 \rightarrow 2}) \leq c \sqrt{s \log (mn)}$, for some constant $c$ \cite{ledo13,vers14} (also see Section~\ref{sec:app}). For the sake of simplicity and to avoid clutter, in the sequel we avoid showing all multiplicative constants (like `$c$' in the last line)
which do not depend on any problem parameters (like sizes of matrices/vectors involved). Since we work with sub-Gaussian random variables, for certain analyses the constants may depend on the $\psi_2$ norm of the sub-Gaussian random variable~\cite{vers18}, but we do not show such constants explicitly. Based on the choice, our results are in fact in order notation, without showing the $O(\cdot)$. This is quite common
in analyses especially based on (generic) chaining, and we are inspired by similar choices in the related literature~\cite{tala14,krmr14}, where the same $c$ is used to denote different constants in an analysis, where the actual constant may keep changing from one line to the next.

Now, by definition of $B_{\cA}(\bxi)$ and based on the decoupling result in Theorem~\ref{th:mds-dec0}, we have
\begin{align}
\| B_{\cA}(\bxi) \|_{L_p} & \leq  \left\| \sup_{A \in \cA} \left| \sum_{j,k=1}^n \xi_j \xi'_k \langle A_j, A_k \rangle \right| \right\|_{L_p} = \left\| \sup_{A \in \cA} \left| \langle A \bxi, A \bxi' \rangle \right| \right\|_{L_p}~.
\label{eq:bp10}
\end{align}
Thus, it suffices for the generic chaining argument to focus on bounding the right hand side of \eqref{eq:bp10}. Details of the analysis are presented in Appendix~\ref{ssec:offd}. The main result for the off-diagonal term is as follows:
\begin{theo}
    Let $\bxi$ be a stochastic process adapted to $F$ satisfying (SP-1) and (SP-2). Then, for all $p \geq 1$, we have 
\begin{equation}
\begin{split}
    \| B_{\cA}(\bxi)\|_{L_p} & \leq \gamma_2(\cA,\| \cdot \|_{2 \rightarrow 2}) \cdot \bigg( \gamma_2(\cA,\| \cdot \|_{2 \rightarrow 2}) + d_F(\cA) \bigg) \\ 
    & \phantom{....} + \sqrt{p} \cdot d_{2 \rightarrow 2}(\cA) \cdot \bigg( \gamma_2(\cA,\| \cdot \|_{2 \rightarrow 2}) + d_F(\cA) \bigg) + p \cdot d_{2 \rightarrow 2}^2 (\cA)~.
\end{split}
\label{eq:offd0}
\end{equation}
\label{theo:offd0}
\end{theo}

%{\bf Bounded random variables.} We consider a special case of interest where the distributions $p(\xi_j|\xi_1,\ldots,\xi_{j-1})$ have bounded support. A bounded support makes both $\xi_j$ and $\xi_j^2$ sub-Gaussian random variables, although with different sub-Gaussian norms~\cite{vers14}. Further, the sequence $\{ \eta_j\}$ with $\eta_j = \xi_j^2 - E|\xi_j|^2$ is also a sub-Gaussian MDS, which simplifies bounding the diagonal term $D_{\cA}(\bxi)$. For this special case, we have the following large deviation bound on $C_{\cA}(\bxi)$:

%\subsection{The Diagonal Term $D_{\cA}(\bxi)$} 
\subsection{The Diagonal Term $D_{\cA}($\texorpdfstring{$\boldsymbol{\xi}$}{\textxi}$)$} 
While the diagonal term 
\begin{equation*}
   D_{\cA}(\bxi)  \triangleq ~\sup_{A \in \cA} \left| \sum_{j=1}^n (|\xi_j|^2 - \E|\xi_j|^2) \| A_j \|_2^2 \right| 
\end{equation*}
does not have any interaction terms of the form $\xi_j \xi_k$, the term depends on centered random variables $|\xi_j|^2 - \E|\xi_j|^2$. Since $\xi_j|f_{1:j}$ is sub-Gaussian, a naive analysis by treating (the centered version of) $|\xi_j|^2|f_{1:j}$ as a sub-exponential random variable~\cite{vers18} will lead to dependencies on the $\gamma_1$ function, yielding an additional multiplicative $\sqrt{\log n}$ term~\cite{sibp15,tala14} on the right hand side of \eqref{eq:offd0} in Theorem~\ref{theo:offd0} corresponding to the off-diagonal terms. Such an analysis will make the diagonal term worse than the off-diagonal term by a factor of $\sqrt{\log n}$. Such an additional multiplicative dependency on $\sqrt{\log n}$ will subsequently lead to worse sample complexity bounds for applications of the result, e.g., an RIP sample complexity of $s (\log p)^2$ \abcomment{please check} for the stochastic process $\bxi$ compared to the well known RIP sample complexity of $s \log p$ for $\bxi$ with i.i.d.~sub-Gaussian entries~\cite{bcfs14,vers18,wain19}.

We bound the diagonal term $D_{\cA}(\bxi)$ using a sharper analysis which avoids the multiplicative $\sqrt{\log n}$ term and in fact exactly matches the bound on the right hand side of \eqref{eq:offd0} in Theorem~\ref{theo:offd0} corresponding to the off-diagonal terms. Our analysis relies on three key  results: symmetrization, de-symmetrization, and contraction~\cite{bolm13,leta91}. While symmetrization is widely used in a variety of analysis~\cite{bame02}, the form of the widely used result relies on the elements of $\bxi$ to be statistically independent~\cite[Lemma 6.3]{leta91}\cite[Theorem 11.4]{bolm13}. Our analysis requires a generalization of the classical result to be able to work with stochastic processes $\bxi$ adapted to $F$ satisfying (SP-1) and (SP-2).
We establish the following generalization of the classical symmetrization result for stochastic processes. We prove the result in Appendix~\ref{ssec:diag}.
\begin{lemm}
Let $\bxi$ be a stochastic process adapted to $F$ satisfying (SP-1) and (SP-2). Let $H: \R_+ \mapsto \R_+$ be a convex function and let $\w=[w_i] \in \R^n$ be any vector such that $H(\sup_{g \in \cG} |w_i g(\xi_i)|) < \infty$ for all $i$. Let $E = \{\eps_i\}$ be a collection of i.i.d.~Rademacher random variables. Then, we have
\begin{equation}
    \E_{\bxi,F}\left[ H\left( \sup_{g \in \cG} \left|  \sum_{i=1}^n w_i \big( g(\xi_i) - \E_{\bxi,F}[g(\xi_i)]   \big) \right| \right)  \right] 
    \leq  \E_{\bxi,F,E} \left[ H \left(  2 \sup_{g \in \cG} \left| \sum_{i=1}^n w_i \eps_i g(\xi_i) \right| \right) \right]~.
\end{equation}
\label{lemm:symm30}
\end{lemm}
The convex functions $H$ for symmetrization in our analysis will be $L_p$ norms for $p \geq 1$.

Existing results on de-symmetrization~\cite[Lemma 6.3]{leta91}\cite[Theorem 11.4]{bolm13} also relies on the elements of $\bxi$ to be statistically independent~\cite[Lemma 6.3]{leta91}\cite[Theorem 11.4]{bolm13}. Our analysis again requires a generalization of the classical result to be able to work with stochastic processes $\bxi$ adapted to $F$ satisfying (SP-1) and (SP-2).
We establish the following generalization of the classical de-symmetrization result for stochastic processes. We prove the result in Appendix~\ref{ssec:diag}.
\begin{lemm}
Let $\bxi$ be a stochastic process adapted to $F$ satisfying (SP-1) and (SP-2). 
Let $H: \R_+ \mapsto \R_+$ be a convex function and let $\w=[w_i] \in \R^n$ be any vector such that $H(\sup_{g \in \cG} |w_i g(\xi_i)|) < \infty$ for all $i$. Let $E = \{\eps_i\}$ be a collection of i.i.d.~Rademacher random variables. Then, we have
\begin{equation}
          \E_{E,F,\bxi} \left[ H \left( \frac{1}{2} \sup_{g \in \cG} \left| \sum_{i=1}^n w_i \epsilon_i (g(\xi_i) - \E_{F,\bxi}[g(\xi_i)] ) \right| \right) \right] 
         \leq 
          \E_{F,\bxi} \left[  H \left( \sup_{g \in \cG} \left|  \sum_{i=1}^n w_i \big( g(\xi_i) - \E_{F,\bxi}[g(\xi_i)]  \big) \right| \right) \right]~.
\end{equation}
\label{lemm:desym20}
\end{lemm}
As in the case of symmetrization, the convex functions $H$ for de-symmetrization in our analysis will be $L_p$ norms for $p \geq 1$.

We also need a specific form of the contraction principle~\cite{leta91,bolm13} for our analysis. In fact, we will directly use the following result~\cite[Lemma 4.6]{leta91} since the result relies on stochastic dominance of marginal distributions but not on statistical independence.
\begin{lemm}
Let $H : \R_+ \mapsto \R_+$ be convex. Let $\{\eta_i\}$ and $\{\gamma_i\}$ be two symmetric sequences of real valued random variables such that for some constant $K \geq 1$ and every $i$ and $t > 0$ we have
\begin{equation}
    P( | \eta_i | > t) \leq K P(|\gamma_i| > t )~.
\end{equation}
Then, for any finite sequence $\{ \x_i \}$ in a Banach space,
\begin{equation}
    \E \left[ H \left( \left\| \sum_i \eta_i \x_i \right\| \right) \right] \leq \E\left[ H\left( K \left\| \sum_i \gamma_i \x_i \right\| \right) \right]~.
\end{equation}
\label{lemm:cont}
\end{lemm}

Our overall approach to bounding the diagonal term $D_{\cA}(\bxi)$ involves using symmetrization, de-symmetrization, and contraction to reduce upper bound on $D_{\cA}(\bxi)$ \qgdelete{with $D_{\cA}(\bxi)$} with $D_{\cA}(\g)$, where $\g$ has i.i.d.~normal entries, and additional terms which can be bounded using generic chaining~\cite{tala14}. Further, being based on i.i.d.~normal entries, $D_{\cA}(\g)$ can be bounded based on existing results~\cite{krmr14}. The reason we can avoid the extra $\sqrt{\log n}$ term from the naive sub-exponential random variable based analysis is that the extra term does not show up for the special case of $\bxi = \g$ due to sharper inequalities possible for the special case of Gaussian~\cite{argi93,pegi99,krmr14}.

Putting everything together we have the following bound on the diagonal term $D_{\cA}(\bxi)$ for stochastic processes $\bxi$ satisfying (SP-1) and (SP-2).
\begin{theo}
Let $\cA \in \R^{m \times n}$ be a collection of $(m \times n)$ matrices. Let $\bxi = \{\xi_i\}$ be a stochastic process adapted to $F=\{F_i\}$ satisfying (SP-1) and (SP-2). Consider the random variable
\begin{equation}
    D_{\cA}(\bxi) = \sup_{A \in \cA} \left| \sum_{j=1}^n ( \xi_j^2 - \E|\xi_j|^2 ) \| A^j \|_2^2 \right|~,
\end{equation}
where $A^j$ denotes the $j^{th}$ column of $A$. Then, we have 
\begin{equation}
\begin{split}
    \| D_{\cA}(\bxi) \|_{L_p} & \leq  
    \gamma_2(\cA, \| \cdot \|_{2 \rightarrow 2}) \big( \gamma_2(\cA, \| \cdot \|_{2 \rightarrow 2}) + d_F(\cA) \big) \\
    & \phantom{\leq} ~+ \sqrt{p} \cdot d_{2 \rightarrow 2}(\cA) \big( \gamma_2(\cA, \| \cdot \|_{2 \rightarrow 2}) + d_F(\cA) \big) + p \cdot d_{2 \rightarrow 2}^2(\cA) ~.
\end{split}
\end{equation}
\label{theo:subg-diag-sharp0}
\end{theo}
Note that the upper bounds for the diagonal term $D_{\cA}(\bxi)$ in Theorem~\ref{theo:subg-diag-sharp0} and the off-diagonal term $B_{\cA}(\bxi)$ in Theorem~\ref{theo:offd0} are the same. We now have the pieces to construct the overall bound.

\subsection{The Main Result} 
Based on the upper bounds on the $L_p$ norms of the off-diagonal and diagonal terms respectively in Theorems \ref{theo:offd0} and \ref{theo:subg-diag-sharp0}, we have our main result as stated below.
\begin{theo}
Let $\cA$ be a set of $(m \times n)$ matrices and let $\bxi$ be a stochastic process adapted to $F$ satisfying (SP-1) and (SP-2). Let
\begin{align}
    M & = \gamma_2(\cA,\| \cdot \|_{2 \rightarrow 2}) \cdot \bigg( \gamma_2(\cA,\| \cdot \|_{2 \rightarrow 2}) + d_F(\cA) \bigg) \\%+  d_F(\cA) \cdot \gamma_2(  \cA, \| \cdot \|_{2 \rightarrow 2} ) \\
    V & = d_{2 \rightarrow 2} (\cA) \cdot \bigg( \gamma_2(\cA,\| \cdot \|_{2 \rightarrow 2}) + d_F(\cA)\bigg) \\
    U & = d_{2 \rightarrow 2}^2 (\cA)~.
\end{align}
Then, for any $\eps > 0$,
\begin{equation}
    P\left(  \sup_{A \in \cA} \left| \|A\bxi\|_2^2 - \E\|A\bxi\|_2^2 \right| \geq c_1 M + \eps \right)
    \leq 2 \exp \left( - c_2 \min \left\{ \frac{\eps^2}{V^2} , \frac{\eps}{U} \right\} \right)~,
\end{equation}
where $c_1, c_2$ are constants which depend on the support.
\label{theo:main1}
\end{theo}
\proof With $C_{\cA}(\bxi), B_{\cA}(\bxi), D_{\cA}(\bxi)$ as defined in \eqref{eq:caxi},\eqref{eq:baxi}, and \eqref{eq:daxi} respectively, we
have
\begin{align*}
    \| C_{\cA}(\bxi)\|_{L_p} &  \leq   \| B_{\cA}(\bxi)\|_{L_p}  +     \| D_{\cA}(\bxi)\|_{L_p} \\
  &  \overset{(a)}{\leq} \gamma_2(\cA, \| \cdot \|_{2 \rightarrow 2}) \big( \gamma_2(\cA, \| \cdot \|_{2 \rightarrow 2}) + d_F(\cA) \big) \\
    & \phantom{\leq} ~+ \sqrt{p} \cdot d_{2 \rightarrow 2}(\cA) \big( \gamma_2(\cA, \| \cdot \|_{2 \rightarrow 2}) + d_F(\cA) \big) + p \cdot d_{2 \rightarrow 2}^2(\cA) ~,
\end{align*}
from Theorems~\ref{theo:offd0} and \ref{theo:subg-diag-sharp0}. With such bounds on the $L_p$ norms of the random variable $C_{\cA}(\bxi)$, the main result follows by using the moment-generating function and Markov's inequality~\cite{bolm13,vers18} as in \eqref{eq:mark1} and \eqref{eq:mark2}. \qed

It is instructive to compare our bounds for stochastic processes $\bxi$ satisfying (SP-1) and (SP-2) to the sharpest existing bound on $C_{\cA}(\bxi)$ for the special case when $\bxi$ has i.i.d.~sub-Gaussian entries~\cite{krmr14}. For this i.i.d.~sub-Gaussian case, \cite{krmr14} showed a large deviation bound based on
\begin{align}
    M' & = \gamma_2(\cA,\| \cdot \|_{2 \rightarrow 2}) \cdot \bigg( \gamma_2(\cA,\| \cdot \|_{2 \rightarrow 2}) + d_F(\cA) \bigg) +  d_F(\cA) \cdot d_{2 \rightarrow 2}(\cA) \\
    V' & = d_{2 \rightarrow 2} (\cA) \cdot \bigg( \gamma_2(\cA,\| \cdot \|_{2 \rightarrow 2}) + d_F(\cA)\bigg) \\
    U' & = d_{2 \rightarrow 2}^2 (\cA) ~.
\end{align}
By comparing the terms with those in Theorem~\ref{theo:main1}, we note that $U = U'$ and $V = V'$ and while $M'$ has an extra additional term
$d_F(\cA) \cdot d_{2 \rightarrow 2}(\cA)$, for symmetric sets $\cA$ with $\cA = -\cA$ we have $d_{2 \rightarrow 2}(\cA) \leq \gamma_2(\cA, \| \cdot \|_{2 \rightarrow 2})$, so the terms are of the same order. Thus, the generalization to the stochastic process $\bxi$ yields the same order bound as the i.i.d.~case which allows seamless extension of applications of the result to random vectors/matrices with statistical dependence~(Section~\ref{sec:app}).

Finally, note that our results can be extended to the case of non-zero mean stochastic processes. In particular with $\x = \bxi + \m$, where $\bxi$ is the stochastic process satisfying (SP-1) and (SP-2) and $\m$ is the mean vector, i.e., $\E[\x] = \m$, we have 
$\|A\x \|^2 - \E\| A \x \|_2^2 = (\|A \bxi \|_2^2 - \E \| A \bxi \|_2^2) + \langle \bxi, 2A^T A \m \rangle$, where the first term is what we analyze and bound in Theorem~\ref{theo:main1}, and the second term is a linear form of $\bxi$. For the unifom bound, the two terms can be separated using Jensen's inequality, the first term can be bounded using Theorem~\ref{theo:main1} and the second term can be bounded using a standard application of generic chaining using (SP-1) and (SP-2). Thus, mean shifted versions of our results also hold.

%\ab{double-check + add discussion} 

%\ab{@Vidyashankar, can you add a brief discussion on this part?}

%\newpage

\section{Implications of the Main Results}
\label{sec:app}
We show several 
%\swdelete{potential} 
applications of our results, including the Johnson Lindenstrauss (J-L), Restricted Isometry Property (RIP), and sketching. 
%In this section, we illustrate implications of the main result to a variety of existing results.

%\qg{I assume the Gaussian width is already defined in previous sections.}

%\qg{Shall we have subsections?}\swcomment{yes, please}\qg{added subsections}
%\vsdelete{\subsection{Johnson-Lindenstrauss with MDS}}
\subsection{Johnson-Lindenstrauss with Stochastic Processes}
\label{subsec:app-jl}
%\qgcomment{Do we still use MDS in the title?}
%When dealing with high-dinamensional data, a widely used approach is to first apply a random projection to embed the data in a low-dimension space while approximately preserving  all pairwise distances in the original high-dimensional data
 Let $X \in \R^{n \times p}$ %\vsdelete{with}
 , $n < p$ and let $\cA$ be any set of $N$ vectors in $\R^p$. $X$ is a {\it Johnson-Lindenstrauss transform} (JLT) \cite{joli84,aich06} if for any $\eps > 0$,
\beq
(1 - \eps)\|u\|_2^2 \leq \|Xu\|_2^2 \leq (1 + \eps)\|u\|_2^2 \quad \text{for all } u \in \cA~.
\label{jl}
\eeq
%\vsdelete{Effectively Johnson-Lindenstrauss} 
JLT is a random projection which embeds high-dimensional data into lower-dimensional space while approximately preserving all pairwise distances \cite{wood14,maho11,inmr97}. JLT has found numerous applications that include 
%bi-Lipschitz embedding of graphs into normed spaces \cite{frma88}, 
searching for an aproximate nearest neighbor in high-dimensional Euclidean space \cite{inmo98}, dimension reduction in data bases \cite{achl03}, learning mixture of Gaussians \cite{dasg99} and sketching \cite{wood14}. It is well known that $X = \frac{1}{\sqrt{n}} \tilde{X}$, where $\tilde{X}$ contains standard i.i.d. normal elements, is a JLT with high probability when $n = \Omega(\log N)$ \cite{joli84}. An immediate consequence of our result is that $X$ with arbitrarily dependent rowwise sequential entries is a JLT with high probability when $n = \Omega(\log N)$.%\qgcomment{I checked again, the sample complexity here is $\log N$, no square root.}
% Effetchinfctively Johnson-Lindenstrauss is a widely used approach when dealing with high-dimensional data is to apply a random projection to embed the data in a lower-dimensional space while approximately preserving all pairwise distances in the original high-dimensional data.  The fundamental result for random dimensionality reduction is the Johnson-Lindenstrauss (J-L) lemma .  %can be viewed as a special case of RIP for finite set and 

%Given a sub-Gaussian MDS sampled the same way as for MDS-RIP, 
%\vsedit{Let $\tilde{X}_{i,j}$ denote the element in the $i$-th row and $j$-th column of $\tilde{X}$ and $\tilde{X}_{i,:}$ denote the $i$-th row of $X$. The entries of $\tilde{X}$ are sequentially generated as follows,}

%\begin{enumerate}
    %\item Initially, the first element of the matrix $X_{1, 1}$ is drawn from a zero-mean distribution.
    %$$E [X_{1, j}| X_{1, j-1}, X_{1, j-2},\ldots,X_{1,1}] = 0$$
%    \item $\tilde{X}_{i+1, j}\mid \{\tilde{X}_{i', :}\}_{i' < i}, \{\tilde{X}_{i, j'}\}_{j' < j}$ is sub-Gaussian 
%    \item $E [\tilde{X}_{i, j}\mid \{\tilde{X}_{i', :}\}_{i' < i}, \{\tilde{X}_{i, j'}\}_{j' < j}] = 0$.
%\end{enumerate}
%\vsedit{It can be easily verified that the data can be modeled as generated from graphical model GM2, where the $\xi$'s are entries of the matrix $\tilde{X}$ generated row wise and the corresponding $F_i$'s are the observed matrix entries generated before $\xi$. Moreover, SP-1 is satisfied from conditions 1 and 2 above and SP-2 is trivially satisfied.}
Let us denote the element in the $i$-th row and $j$-th column of $\tilde{X}$ as $\tilde{X}_{i, j}$, and the $i$-th row as $\tilde{X}_{i, :}$. 
%\vsdelete{We consider a random matrix $X$ such that $\text{vec}(\tilde{X})$ is sub-Gaussian MDS:}. 
Let the entries of $\tilde{X}_{i,j}$ being sequentially generated as follows:

%\vsdelete{
%\begin{enumerate}
%    \item Initially, the first element of the matrix $X_{1, 1}$ is drawn from a zero-mean distribution.
%    $$E [X_{1, j}| X_{1, j-1}, X_{1, j-2},\ldots,X_{1,1}] = 0$$
%    \item $X_{i+1, j}\mid \{X_{i', :}\}_{i' < i}, \{X_{i, j'}\}_{j' < j}$ is sub-Gaussian 
%   \item $E [X_{i, j}\mid \{X_{i', :}\}_{i' < i}, \{X_{i, j'}\}_{j' < j}] = 0$.
%\end{enumerate}
%}
%\vsedit{
\begin{enumerate}
    \item Initially, the first element of the matrix $\tilde{X}_{1, 1}$ is drawn from a zero-mean sub-Gaussian distribution.
    \item $\tilde{X}_{i,j}$ is a conditionally 1-sub-Gaussian random variable satisfying $\E[\tilde{X}_{i,j} | f_{i,j}] = 0$. The $f_{i,j}$ are realizations of a stochastic process which can possibly depend on the entries $\{ \{ \tilde{X}_{i',:} \}_{i' < i}, \{ \tilde{X}_{i,j'} \}_{j' < j} \}$, i.e., the elements in the previous rows and columns to the element $(i,j)$.
    \item $\tilde{X}_{i,j} \perp \{ \{ \tilde{X}_{i',:} \}_{i' < i}, \{ \tilde{X}_{i,j'} \}_{j' < j} \} ~ | ~ f_{i,j}$ and $\tilde{X}_{i,j} \perp \{ \{f_{i,j'} \}_{j' > j},\{f_{i',:} \}_{i' > i} \} ~ | ~ f_{i,j} $
\end{enumerate}
%}
%Given a sub-Gaussian MDS sampled from the above process, %we show a sample complexity bound for obtaining the RIP in $X$. 
%\vs{Not clear need a reference to the earlier explanation}, 
%The following conclusion showing $X$ is a JLT with high probability follows from the result of Theorem \ref{theo:main2}:
\begin{corollary}[JL]
    Let $X \in \R^{n \times p}$ be a matrix constructed as $X = \frac{1}{\sqrt{n}}\tilde{X}$. %\vsdelete{is a bounded MDS}
    If we choose $n = \Omega(\epsilon^{-2}\log N)$, $X$ is a JLT with probability at least $1 - \frac{1}{N^c}$ for a constant $c > 0$. 
    \label{lemm:jl}
\end{corollary}

\proof
    To make use of Theorem~\ref{theo:main1}, %let us first rearrange $X$ to a vector by concatenating the rows of $X$. We denote the rearranged vector as $\text{vec}(X)$ and
    %We formulate each entry of $X$ by $X =  \frac{1}{\sqrt{n}} S$. We assume that $\text{vec}(S)$ is an $L$-sub Gaussian unit variance MDS. 
    let $x = [\tilde{X}_{1, :}, \tilde{X}_{2, :}, \ldots, \tilde{X}_{n, :} ]^T$ be a sub-Gaussian random vector of length $np$ by concatenating the rows of $\tilde{X}$. We can rewrite $\frac{1}{\sqrt{n}}\tilde{X} \theta = V_{\theta} x$, 
    where $V_{\theta} \in \R^{n \times np}$ is a block diagonal matrix
    \beq
        V_{\theta} = \frac{1}{\sqrt{n}}\begin{bmatrix}
            \theta^T & 0 & \cdots & 0 \\
            0 & \theta^T & \cdots & 0 \\
            \vdots & \vdots & \ddots & \vdots \\
            0 & 0 & \cdots & \theta^T
        \end{bmatrix}.
        \label{eq:vtrans}
    \eeq
     Let $\cA$ be any set of $N$ unit vectors in $\R^p$, then \eqref{jl} is equivalent to 
     \[
        1 - \eps \leq \|X u\|_2^2 \leq 1 + \eps\quad\text{for all}~ u \in \cA.
     \]
     In this case, we have $\|V_\theta\|_F = \|\theta\|_2 = 1$. Therefore $d_F(\cA) = 1$. Besides, we have $\|V_\theta\|_{2 \rightarrow 2} = \frac{1}{\sqrt{n}}\|\theta\|_2 = \frac{1}{\sqrt{n}}$ and $d_{2 \rightarrow 2}(\cA) = \frac{1}{\sqrt{n}}$., from results in \cite{tala14,vers12}, there is constant $C > 0$ such that
    \beq
        \gamma_2(\cA, \|\cdot\|_{2 \rightarrow 2}) \leq C \frac{w(\cA)}{\sqrt{n}} \leq C \sqrt{\frac{ \log N}{n}}.
        \label{bnd:gamma}
    \eeq
    Therefore we have
    \beq
        M_1 = \calO \left( \frac{\log N}{n} + 2 \sqrt{ \frac{\log N}{n}} \right),
        ~~ V = \calO \left( \frac{1}{\sqrt{n}}\right),
        %\left( \sqrt{\frac{s \log (2p / s)}{n}} + 1 \right) \right),
    ~\text{and}~
        U_1 = \frac{1}{n}.
        \label{bnd:mvu}
    \eeq
    Thus, when $n = \Omega (\eps^{-2} \log N )$, $\max_i |\|X \theta_i\|_2^2 - \|\theta_i\|_2^2| \leq \eps$ %\vsdelete{$n = \Omega (\eps^{-2} \log N )$, $\max_i |\frac{1}{n}\|X \theta_i\|_2^2 - \|\theta_i\|_2^2| \leq \eps$} 
    with probability at least $1 - e^{-c \log N}$ for constant $c > 0$.
\qed

%We can get Lemma \ref{lemm:jl} for MDS $X$ similar to MDS-RIP since the Gaussian width of $N$ unit vectors is $\calO(\log N)$. 
%\vsdelete{Traditionally the result of the J-L lemma assumes each entry of $X$ is sampled i.i.d.~from a standard normal distribution, then $X$ is scaled with $\frac{1}{\sqrt{n}}$. We show that the result of the J-L lemma holds for sub-Gaussian MDS $X$ even with there are arbitrary dependencies between the entries of $X$.  Our result for $n$ matches the result for the i.i.d.~case} 
%even though there are statistical dependencies between entries of the matrix.

%We name our result MDSJL.

%\begin{lemma}[MDSJL]
%    Let $\theta_1, \theta_2, \ldots, \theta_N \in \R^p$ be arbitrary points such that $\|\theta_i\|_2 = 1$. For any $\eps > 0$, as long as $n = \calO (\eps^{-2} \log N)$, there exists a matrix $X \in \R^{n \times p}$ such that
%    \beq
%        (1 - \eps ) \leq \|X \theta_i\|_2^2 \leq (1 + \eps)   \quad \text{for all }i =  1, 2, \ldots, N.
%    \eeq
%    \label{lemm:mdsjl}
%\end{lemma}

\subsection{Restricted Isometry Property (RIP) with Stochastic Processes}
Matrices satisfying Restricted Isometry Property (RIP) are approximately orthonormal on sparse vectors \cite{cata05,cata07}. Let $X \in \R^{n \times p}$ and let $\cA$ be the set of all $s$-sparse vectors in $\R^p$. We define matrix $X$ to satisfy RIP with the restricted isometry constant $\delta_s \in (0,1)$ if for all $u \in \cA$,
\beq
(1 - \delta_s)\|u\|_2^2 \leq \frac{1}{n} \|X u\|_2^2
%\underset{u \in \cA}{\inf} \frac{1}{n} \|Xu\|_2^2 \leq \underset{u \in \cA}{\sup} \frac{1}{n} \|Xu\|_2^2 
\leq (1 + \delta_s)\|u\|_2^2 ~.
\eeq
Matrices satisfying RIP are of interest in high-dimensional statistics and compressed sensing problems where the goal is to recover a sparse signal $\theta^* \in \R^p$ from limited noisy linear measurements. Formally, given data $(X,y)$ assumed to be generated according to the linear model $y = X\theta^* + \omega$, with $y \in \R^{n}$, $X \in \R^{n \times p}$, $\theta^* \in \R^p$ is $s$-sparse, $\omega \in \R^n$ is the unknown noise, the goal is to obtain good estimates of $\theta^*$ when $n,s << p$. This is achieved with the Lasso estimator $\hat{\theta} = \underset{\theta}{\argmin} \frac{1}{2n} \|y - X\theta\|_2^2 + \lambda_n \|\theta\|_1$ \cite{tibs96}. A sufficient condition for Lasso to work, is that the design matrix should satisfy RIP \cite{cata05,cata07}. It has now been established that sub-Gaussian random matrices with i.i.d. rows, e.g., rows sampled from a $N(0,\sigma^2 \I_{p \times p})$ satisfies RIP \cite{cata05,cata07,nrwy12,bcfs14} when $n = \Omega(s \log p)$. But the i.i.d. rows assumption is violated in many practical settings when data is generated adaptively/sequentially. Examples include times-series regression and  bandits problems \cite{lutk05, kmrw18}, active learning \cite{sett12,hann14} or volume sampling \cite{dewa18,dewh18}. An application of our new results shows that the i.i.d. assumption is not necessary and design matrices generated from dependent elements also satisfy RIP when $n = \Omega(s \log p)$.
\paragraph{RIP for sub-Gaussian designs with dependent entries.}
We consider matrices $X$ generated as matrix $\tilde{X}$ in Section \ref{subsec:app-jl}. To recap the entries of the design matrix satisfy the following properties,
\begin{enumerate}
    \item Initially, the first element of the matrix $X_{1, 1}$ is drawn from a zero-mean sub-Gaussian distribution.
    \item $X_{i,j}$ is a conditionally 1-sub-Gaussian random variable satisfying $\E[X_{i,j} | f_{i,j}] = 0$. The $f_{i,j}$ are realizations of a stochastic process which can possibly depend on the entries $\{ \{ X_{i',:} \}_{i' < i}, \{ X_{i,j'} \}_{j' < j} \}$, i.e., the elements in the previous rows and columns to the element $(i,j)$.
    \item $X_{i,j} \perp \{ \{ X_{i',:} \}_{i' < i}, \{ X_{i,j'} \}_{j' < j} \} ~ | ~ f_{i,j}$ and $X_{i,j} \perp \{ \{f_{i,j'} \}_{j' > j},\{f_{i',:} \}_{i' > i} \} ~ | ~ f_{i,j} $
\end{enumerate}
\begin{corollary}[RIP]
    \label{cor:mdsrip}
    Let $X \in \R^{n \times p}$ be a matrix generated from the above process.
    Then %for an arbitrary set ~$\Omega \subset \R^p$, 
    for any $\eps > 0$, if we choose $n = \Omega(\eps^{-2} s \log (2p / s))$,  then $\delta_s \leq \eps$ with probability at least $1 - \left( \frac{s}{2p} \right)^{cs}$ for a constant $c > 0$.
    \label{coro:rip_mds_sparse}
\end{corollary}

The result can be shown using similar arguments as for JL noting that the Gaussian width of unit $s$-sparse vectors is $\calO(s\log (2p/s))$ \cite{crpw12}.
%\vsdelete{We can get Corollary \ref{cor:mdsrip} for $X$ similar to $\tilde{X}$ in JL since the Gaussian width of unit $s$-sparse vectors is $\calO(s\log (2p/s))$ \cite{crpw12}.}

%\vscomment{Can we delete the below paragraph? I haved moved some parts of this to the top.}\qgcomment{Shall we talk about estimation from general structures other than sparse? If not let's remove it.}

%As a concrete example, we can apply our result to the following adaptive linear regression setting in which labelled data $(\x_i, y_i)$ arrives sequentially such that each pair is generated from a noisy linear model $y_i = \langle \x_i,\theta^*\rangle + \eps$, where $\theta^*$ has structure like sparsity or low-rank characterized by a norm $R(\cdot)$, so $R(\theta) \leq R(\theta^*)$~\cite{crpw12,bcfs14} and the sequence of $\x_i$'s are chosen (element-wise and row-wise) adaptively as a sub-Gaussian MDS (e.g., by a bandit or active learning algorithm). While prior work~\cite{NTTZ18,DMST18,NR18} shows that such adaptivity can induce bias in the estimate $\hat\theta$ obtained using the design $X$ where the rows $\xi_i$ are dependent, our result provides a tool to give a high-probability bound on $\|\hat \theta - \theta\|_2$ that is obtained via norm regularized regression \cite{bcfs14,crpw12}. 
%\beq
 %   \min_\theta \frac{1}{n}\|y - X\theta\|_2^2 + \lambda_n R(\theta).
 %   \label{eq:reg}
%\eeq
%where $R(.)$ is a norm characterizing the structure of $\theta^*$.

\paragraph{RIP for partial Toeplitz matrix designs.} %Another potential application of Theorem~\ref{theo:main1} is for showing RIP of partial Toeplitz matrices \cite{rauh09,bahr07}. 
Let $\bxi \in \R^{2p - 1} = (\xi_j)$. A matrix $A_{\bxi}$ is called a Toeplitz matrix if it has the following form
\beq
    A_{\bxi} = \begin{bmatrix}
        \xi_p & \xi_{p + 1} & \cdots & \xi_{2 p - 1} \\
        \xi_{p - 1} & \xi_p & \cdots & \xi_{2 p - 2} \\
        \vdots & \vdots & \ddots & \vdots \\
        \xi_1 & \xi_2 & \cdots & \xi_p.
    \end{bmatrix}
\eeq
Toeplitz matrices are widely used in compressed sensing and are desirable alternatives for random matrices with i.i.d. entries because (1) we only need to maintain $2p - 1$ random variables and (2) multiplication with a Toeplitz design can be efficiently implemented using fast Fourier transform (FFT)~\cite{habr10,duel11}.
Let $R \in \R^{n \times p}$ be a (deterministic) matrix that selects $n$ elements of a vector in $\R^p$ \cite{rart10}. %and scales each element by $\frac{1}{\sqrt{n}}$. 
The matrix $X = R \cdot A_{\bxi}$ is used as a sensing matrix. \cite{krmr14} showed that $X$ satisfies RIP when $\bxi$ is sampled i.i.d.~from a univariate sub-Gaussian distribution. We show that RIP holds for Toeplitz matrices even when the elements of $\bxi$ are generated as follows.
\begin{enumerate}
    \item $\bxi_i$ is 1-sub-Gaussian depends on realizations  $f_i$ of a stochastic process which can depend on $\bxi_{1},\bxi_{2},\hdots,\bxi_{i-1}$ and $\E[\bxi_i | f_i] = 0$.
    \item $\bxi_i \perp \bxi_{i'} ~ | ~ f_i,~ i' < i$ and $\bxi_i \perp f_k ~|~ f_i, ~ k > i$.
\end{enumerate}
%\vscomment{Check if this is true. Also can the matrix $R$ be adaptive.}
%\qgcomment{$R$ can be adaptive}
%Based on Theorem~\ref{theo:main1}, we have the following RIP for Toeplitz matrices.

\begin{corollary}[Toeplitz RIP]
    Suppose $\bxi \in \R^{2 p - 1}$ 
    and $X = R A_{\bxi}$ is a partial Toeplitz matrix. Then, 
%    \ab{how is $\Omega$ related to other variables or the result being presented? how can it be arbitrary?}\qg{removed $\Omega$} 
    if we choose $n = \Omega (\eps^{-2}s \log (2p / s))$, RIP 
    %\ab{what is generalized RIP? has it been defined?} 
    holds for $X$ with probability at least $1 - \left( \frac{s}{2p} \right)^{cs}$ with constant $c > 0$.
\end{corollary}
\proof
Let
\beq
    V_{\theta} = \frac{1}{\sqrt{n}}\begin{bmatrix}
        \theta_1 & \theta_2 & \theta_3 & \cdots & 0 & 0 \\
        0 & \theta_1 & \theta_2 & \cdots & 0 & 0 \\
        0 & 0 & \theta_1 & \cdots & 0 & 0 \\
        \vdots & \vdots & \vdots & \ddots & \vdots & \vdots \\
        0 & 0 & 0 & \cdots & \theta_{p - 1} & \theta_p
    \end{bmatrix},
\eeq
we have $\frac{1}{\sqrt{n}}A_{\bxi} \theta = R_{\Omega} V_\theta \bxi $. Let $\cA$ be the 
set of all unit $s$-sparse vectors in $\R^p$. The proof is similar to that for Corollary \ref{cor:mdsrip}. 
%\ab{please provide a bit more details.} 
We have $\|R_{\Omega} V_\theta\|_F = \|\theta\|_2$ and $\|R_{\Omega} V_\theta\|_{2 \rightarrow 2} \leq \frac{1}{\sqrt{n}}\|\theta\|_2$. Then we can bound $\gamma_2(\cA, \|\cdot\|_{2 \rightarrow 2})$, $M_1$, $V$, and $U_1$ similarly as \eqref{bnd:gamma} and \eqref{bnd:mvu}. \qed

\iffalse
\qg{I am not very familiar with literature on Toeplitz compressed sensing. Can I use bound on random circulant matrices since its a special case?}
\swcomment{but we didn't say what's the purpose for having RIP here...} \qg{explained: Toeplitz matrix can be more efficiently computed and in some cases we can only use a Toeplitz design}
\fi
%{\bf Example: Regularized regression}

\paragraph{RIP for adaptively generated rows.}

Sequential learning problems like linear contextual bandits involve estimating a parameter with a design matrix whose rows are adverserially generated based on previously observed rows and rewards which are linear functions of the rows. An example is the linear contextual bandit problem considered, e.g., in \cite{kmrw18,siwb19}. The data generation in \cite{kmrw18,siwb19} can be modeled with graphical model GM3.
%Consider the following data generation process which can be modeled with graphical model GM3.
\begin{enumerate}
\item Let $\mathcal{H}_{t-1}$ denote historical data observed until time $t-1$. In time step $t-1$ an adaptive adversary $\mathcal{A}_{t-1}$  maps the histories to $k$ contexts $\mu_t^1,\hdots,\mu^k_t$ in $\R^p$ with $\|\mu_t^1\|_2 \leq 1$, i.e., $\mathcal{A}_{t-1}: \mathcal{H}_{t-1} \rightarrow (B_2^p)^k$ where $B_2^p$ represents the unit ball in $p$ dimensions. Nature perturbs the contexts with random Gaussian noise, i.e., $x_t^i = \mu_t^i + g_t^i$ with $g_t^i \sim N(0,\sigma^2 \I_{p \times p})$. Now, in the context of GM3, $\mathcal{H}_{t-1} \cup \{x_t^1,\hdots,x_t^k \}$ represents $F_{1:t-1}$. 
\item In time step $t$, a learner chooses one among $k$ contexts $\{x_t^1,\hdots,x_t^k \}$ based on historical data $\mathcal{H}_{t-1}$. Let $x_t^{i_t}$ denote the selected context and $g_t^{i_t}$ denote the corresponding Gaussian perturbation. In the context of GM3, we denote the centered Gaussian perturbation $g_t^{i_t} - \E[g_t^{i_t}]$ by $\xi_t$. %\abcomment{not sure I follow, $x_t^{i_t}$ is a deterministic function of $F_{1:(t-1)}$ (that is ok), but is {\em not} zero mean -- do we not need a conditional centering to get $\xi_t$?}\vscomment{Take care of this later in the exposition}
The learner receives the noisy reward $y_t = \langle x_t^{i_t},\theta^* \rangle + \omega_t$ where $\omega_t$ is an unknown sub-Gaussian noise. History at time step $t$ is now augmented with the new data, i.e., $\mathcal{H}_t = \mathcal{H}_{t-1} \cup \{ \{x_t^1,\hdots,x_t^k \}, x_t^{i_t},y_t \}$.
\item Now similar to step 1, the contexts in time step $t$, $\{x_{t+1}^1,\hdots,x_{t+1}^k \}$, are generated by an adversary $\mathcal{A}_t: \mathcal{H}_t \rightarrow (B_2^p)^k$ perturbed with Gaussian noise and $\mathcal{H}_t \cup \{x^1_{t+1},\hdots,x^k_{t+1} \}$ represents $F_{1:t}$.
%\abcomment{are these contexts the means of the context random vectors? that is what the smoothed model assumes} 
%\item A learner chooses a context $\xi_t \in \R^p$ from among the $k$ contexts using data on previously chosen contexts $\xi_1,\hdots,\xi_{t-1}$ \abcomment{if $\xi_t$ is a direct function of $\xi_{t-1}$, then (SP-2) is not satisfied; $\xi_t$ needs to be a function of $F_{t-1}$.} and noisy linear reward functions $y_i = \langle \xi_i,\theta^* \rangle + \omega_i, ~ 1 \leq i \leq t-1$ and receives reward $y_t = \langle \xi_t,\theta^* \rangle + \omega_t$. 
%\item The domain for the stochastic process \abcomment{this is almost a measurability statement, i.e., domain of a random variable} $F_t$ in time step $t$ is $\mathcal{H}_{t-1} \cup \{\{x^1_t,\hdots,x^k_t \},\xi_t,y_t \}$.
\end{enumerate}
The data generation process mirrors GM3 with $F_{t}$ being a sub-Gaussian process which is influenced by $F_{t-1}$ and $\xi_t$ but generated adaptively by an adversary. \abdelete{instead of any well-defined function\abcomment{not sure i follow -- do we need the last bit?}.} $\xi_t$ is a sub-Gaussian random vector chosen by the learner using historical data $\mathcal{H}_{t-1}$ satisfying (SP-2). Specifically in \cite{siwb19,kmrw18}, the parameter $\hat{\theta}$ is estimated using the least squares estimator with contexts $x_t^{i_1},\hdots,x_{t-1}^{i_{t-1}}$ observed in the $(t-1)$ previous time steps, response as the corresponding rewards $y_1,\hdots,y_{t-1}$ and $x_t^{i_t}$ is chosen greedily, i.e, $x_t^{i_t} = \underset{x_t^i:1 \leq i \leq t}{\argmax} \langle x_t^i,\hat{\theta} \rangle$. Let $X_t$ denote the centered design matrix which has the rows $\xi_1,\hdots,\xi_{t-1}$.
%Let $X_t$ denote the design matrix which has centered rows \abcomment{centering needs to be conditioned on the realization so far?} $\xi_i - E[\xi_i], 1 \leq i \leq t$.  
%Note that the centered rows can be computed using empirical expectations \cite{kmrw18,siwb19} \vscomment{Check this. Should we add a citation to the structured bandits paper mentioning it is under preparation?}.  \abcomment{yes, please cite Sivakumar-Wu-Banerjee'19 in prep}
%In many sequential learning problems, the learner estimates the parameter $\theta^*$ using the least squares estimator with design matrix $X_t$ containing rows as $\xi_1,\hdots,\xi_{t-1}$ and response the rewards $y_1,\hdots,y_{t-1}$. 
A critical condition in the analysis for efficient estimation of the parameter in time step $t$ requires the design matrix $X_t$ to satisfy non-asymptotic lower bounds of the RIP condition for some positive constant $\epsilon$,
\beq
\left ( \inf\limits_{u \in \R^p}\E[\|X_{t}u\|_2^2] - \epsilon \right ) \|u\|_2^2 \leq \inf\limits_{u \in \R^p} \|X_{t} u\|_2^2 ~.
\eeq
assuming the result holds in expectation. \abcomment{are we saying our results help move from the in-expectation result to the high-probability result?}\vscomment{Yes. The argument in the traditional bandit setting without the smoothing is different and does not use the minimum eigenvalue condition.}. It was shown in \cite{kmrw18,siwb19} that $\inf\limits_{u \in \R^p} \E[\|X_t u\|_2^2] \geq t \kappa \|u\|
_2^2$, \swcomment{min over unit sphere?} where $\kappa$ is a constant whose value depends on the problem parameters like the number of contexts $k$ and the total number of rounds $T$. The following non-asymptotic results follows directly from our result.
\begin{corollary}
Let $X_t$ be a design matrix generated from the process described above. % with the condition that $E[\xi_t | F_{t-1}] = 0$. 
Then for any $\epsilon > 0$, if we choose $t = \Omega(\epsilon^{-2} \kappa^{-2} p) $, then with probability atleast $1 - \exp(-c p)$ for constant $c > 0$, the following condition is satisfied,
\beq
\inf\limits_{u \in \R^p} \|X^t u\|_2^2 \geq t \kappa (1 - \epsilon) \|u\|_2^2 ~.
\eeq
\end{corollary}
\proof The result follows directly from similar arguments as the result of Corollary \ref{cor:mdsrip} by noting that for the least squares estimator $\cA = S^{p-1}$ and $\gamma_2(\cA,\|\cdot\|_{2 \rightarrow 2}) \leq C \sqrt{\frac{p}{t}}$.

\subsection{CountSketch}
%In this example, we look at J-L for the CountSketch matrix \cite{chcf02,wood14}. 
CountSketch or sparse JL transform is used in real world applications like data streaming and dimensionality reduction \cite{chcf02,wood14}. Every column of a $(n \times p)$ CountSketch matrix $X$ has only $d (d \ll n)$ non-zero elements, therefore for any vector $u \in \R^p$, computing $X u$ takes only $O(d p)$ instead of $O(np)$. Each entry of a CountSketch matrix $X$ is given by $X_{i, j} = \eta_{i, j} \delta_{i, j} / \sqrt{d}$, where $\delta_{i, j}$ is an independent Rademacher random variable, and $\eta_{i, j}$ is a random variable sampled adaptively.
%, and $d$ is the sparsity level of a column. 
The $\eta_{i, j}$ satisfy
%\[
$\sum_{i=1}^n \eta_{i, j} = d,~ \eta_{i, j} \in \{0, 1\}$,
%\]
that is each column has exactly $d$ non zero elements. 
For every column $j$ of $X$, the $\eta_{i,j}$ can be generated by sampling $d$ indices from $\{1, 2, \ldots, n\}$ adaptively given previous columns, then set corresponding $X_{i, j}$ to be a Rademacher random variable, so that $X_{i,j}$ depends on $X_{1,j}, X_{2, j}\ldots, X_{i-1, j}$. %$\text{vec}(X)$ forms a MDS since the independent $\delta_{i,j}$ has mean zero. 
The data generation process of countSketch matrix follows graphical model GM1.
%\ab{how? please show this using a simple argument} 
 %Thus, Corollary \ref{lemm:jl} provides a powerful tool to design and analyze dimensionality reduction algorithms like CountSketch.
The variance of $X_{i,j}$ is $\frac{1}{n}$ and since all the entries of $X$ are bounded by $1$, $X$ is a JLT over $N$ points when the number of rows satisfies $n = \Omega (\epsilon^{-2} \log N )$.
%\vsdelete{The variance of $X_{i,j}$ is $\frac{1}{n}$. Therefore $X$ is a JLT  
%For $N$ points, since all the entries of $X$ are bounded by $1$,
%by Corollary \ref{lemm:jl} we can have $n = \Omega (\epsilon^{-2} \log N )$ rows   
%\ab{what is $N$ in the streaming setting? how do they choose $d$ in the streaming setting? can we keep increasing $d$ as more samples come in? i dont know much about this -- but try to help the reader (like me)} 
%such that $X$ is a JLT.}
Unlike \cite{daks10,kane14}, our bound does not depend on the choice of $d$. Our bound also matches the state of the art \cite{kane14}.

%\qg{We probability don't have time for a discussion on approximate linear algebra.}
%\subsection{Sketching with MDS}

%\subsection{Approximate Linear Algebra with MDS}

% \section{Analysis based on Generic Chaining}
% \label{sec:sketch}
% \input{sec/sketch}

\section{Conclusions}
\label{sec:conc}

Several existing results in machine learning and randomized algorithms, e.g., RIP, J-L, sketching, etc., rely on uniform large deviation bounds of random quadratic forms based on random vectors or matrices. Such results are uniform over suitable sets of matrices or vectors, and have found wide ranging applications over the past few decades. Growing interest in adaptive data collection, modeling, and estimation in modern machine learning is revealing a key limitation of such results: the need for statistical independence, e.g., elementwise independence of random vectors, row-wise independence of random matrices, etc. In this paper, we have presented a generalization of such results \qgdelete{to work with sub-Gaussian MDSs, which} \qgedit{that} allows for statistical dependence on the history. We have also given examples for certain cases of interest, including RIP, J-L, and sketching, illustrating that in spite of allowing for dependence, our bounds are of the same order as that for the case of independent random vectors. We anticipate our results to simplify and help make advances in analyzing learning settings based on adaptive data collection. Further, the added flexibility of designing random matrices sequentially may lead to computationally and/or statistically efficient random projection based algorithms. In future work, we plan to sharpen our analysis especially for the case of unbounded sub-Gaussian random variables and also investigate applications of these results in adaptive data collection and modeling settings.

{\bf Acknowledgements:} The research was supported by NSF grants OAC-1934634, IIS-1908104, IIS-1563950, IIS-1447566, IIS-1447574, IIS-1422557, CCF-1451986, a Google Faculty Research Award, a J.P. Morgan Faculty Award, and a Mozilla research grant. Part of this work completed while ZSW was visiting the Simons Institute for the Theory of Computing at UC Berkeley.  
%\newpage

\bibliographystyle{plain}
\bibliography{supp_norm_ref}

%\newpage

\appendix

\section{Decoupling for Stochastic Processes}
\label{sec:mart-dec}

% \begin{figure}[t]
% \centering
%  \includegraphics[width = 0.6 \textwidth]{figs/mds_ci.pdf}
%  \caption[]{Graphical Model 1 (GM1) structure for stochastic processes $\{\xi_i\}$ and $\{ F_i\}$ which satisfies (SP-2) by construction (Proposition~\ref{prop:gm1}). While we show arrows only from one random variable, e.g., $F_{i-1} \rightarrow \xi_i$, the process $\xi_i|F_{1:(i-1)}$ implies (potential) dependence on the entire history $F_{1:(i-1)}$. All these arrows are not depicted in this and other figures to avoid clutter.}
% \label{fig:ci}
% \end{figure}

In this section, we establish our main decoupling result.
We also briefly discuss motivations behind our assumptions by revisiting classical notions of decoupled tangent sequences for stochastic processes.

\subsection{Main Decoupling Result}

We state our main decoupling result below:
\begin{theo}
Let $\bxi = \{\xi_i\}$ be a stochastic process adapted to $F=\{ F_i \}$ satisfying (SP-1) and (SP-2). Let $\bxi' = \{ \xi'_i \}$ be any decoupled tangent sequence to $\bxi = \{ \xi_i \}$ so that (DTS-1) and (DTS-2) are satisfied.
%$\P(\xi_i | F_{1:(i-1)}) = \P(\xi'_i | F_{1:(i-1)})$ and $\xi_i \perp \xi'_i \mid F_{1:(i-1)}$ (Figure~\ref{fig:tgtci}).
%, with $\{ \xi'\}$  being conditionally independent given $\cG = \sigma(\{d_i\})$.
Let ${\cal B}$ be a collection of $(n \times n)$ symmetric matrices. Let $h: \R \mapsto \R$ be a convex function. Then,
\beq
\E_{\bxi} \left[ \sup_{B \in {\cal B}} h \left( \sum_{\substack{j,k=1\\ j\neq k}}^n \xi_j \xi_k B_{j,k} \right) \right]
\leq 4 E_{\bxi,\bxi'}\left[ \sup_{B \in {\cal B}} h \left( \sum_{\substack{j,k=1}}^n \xi_j \xi'_k B_{j,k} \right) \right]~.
\eeq
\label{th:mds-dec}
\end{theo}

Our proof uses the following result characterizing distributional equivalence of quadratic forms of DTSs. 
%Note that our main result needs decoupled tangent sequences where the additional decoupling property will be used to handle the diagonal terms. We start with the following result:
\begin{prop}
Let $\Xi = \{\xi_i\}$ be a stochastic process adapted to $\{ {\cal F}_i \}$ satisfying (SP-1) and (SP-2). Let $\Xi' = \{ \xi'_i \}$ be any decoupled tangent sequence to $\Xi = \{ \xi_i \}$ . Let $\tilde{B}$ be a symmetric $(n \times n)$ matrix. Consider the random variables
\begin{align}
    X_n = \sum_{\substack{j,k=1\\ j \neq k}}^n \xi_j \xi_k \tilde{B}_{j,k}~, \qquad \text{and} \qquad X'_n = \sum_{\substack{j,k=1 \\ j \neq k}}^n \xi_j \xi'_k \tilde{B}_{j,k}~.
\end{align}
Then $X_n$ and $X'_n$ are identically distributed. 
\label{prop:dist1}
\end{prop}
\proof For any realization $f_{0:n}$ of $F_{0:n}$, for any $\tilde{B}$, the conditional distributions of $X_n | f_{0:n}$ and $X'_n | f_{0:n}$ are identical, i.e.,
\begin{equation*}
    \P(X_n \leq x \mid f_{0:n}) =  \P(X'_n \leq x \mid f_{0:n})~, \qquad \forall f_{0:n}~,
\end{equation*}
since for all $k$ and $j \neq k$, $\xi_k \perp \xi_j | f_{0:n}$, $\xi'_k \perp \xi_j | f_{0:n}$, and conditioned on $f_{0:n}$, $\xi_k$ and $\xi'_k$ are identically distributed. As a result, we have
\begin{align*}
    \int_{f_{0:n}} \P(X_n \leq x \mid f_{0:n}) p(f_{0:n}) df_{0:n} & =    \int_{f_{0:n}} \P(X'_n \leq x \mid f_{0:n}) p(f_{0:n}) df_{0:n} \\
    \Rightarrow \qquad \P(X_n \leq x) & = \P( X'_n \leq x)~.
\end{align*}
That completes the proof. \qed

{\em Proof of Theorem~\ref{th:mds-dec}:} Let $\Delta = \{\delta_1,\ldots,\delta_n\}$ be a set of i.i.d.~Bernoulli random variables with $P(\delta_i=0) = P(\delta_i=1) = 1/2$. Since $B \in {\cal B}$ are symmetric, we have 
\beq
\sum_{\substack{j,k=1\\ j\neq k}}^n \xi_j \xi_k B_{j,k} 
%= 2 \sum_{\substack{j,k=1\\ j < k}}^n \xi_j \xi_k B_{j,k}
= 4 \E_{\Delta}\left[ \sum_{\substack{j,k=1\\ j\neq k}}^n \delta_j (1-\delta_k) \xi_j \xi_k B_{j,k} \right]~.
\eeq
Since $h : \R \mapsto \R$ is a convex function, by Jensen's inequality
\begin{align*}
h \left( \sum_{\substack{j,k=1\\ j\neq k}}^n \xi_j \xi_k B_{j,k} \right)
& =   h \left( 4 \E_{\Delta}\left[ \sum_{\substack{j,k=1\\ j \neq k}}^n \delta_j (1-\delta_k) \xi_j \xi_k B_{j,k} \right] \right) \\
& \leq 4 \E_{\Delta} h \left( \sum_{\substack{j,k=1\\ j \neq k}}^n \delta_j (1-\delta_k) \xi_j \xi_k B_{j,k} \right)~ \\
\Rightarrow \quad \sup_{B \in {\cal B}} h \left( \sum_{\substack{j,k=1\\ j\neq k}}^n \xi_j \xi_k B_{j,k} \right)
& \leq 4 \sup_{B \in {\cal B}} \E_{\Delta} h \left(  \sum_{\substack{j,k=1\\ j \neq k}}^n \delta_j (1-\delta_k) \xi_j \xi_k B_{j,k} \right) \\
%& \leq E_{\Delta} \left[ \sup_{B \in {\cal B}}  F \left( 4 \sum_{\substack{j,k=1\\ j\neq k}}^n \delta_i (1-\delta_j) \xi_j \xi_k B_{j,k} \right) \right] \\
\Rightarrow \quad \E_{\Xi | F} \left[ \sup_{B \in {\cal B}} h \left( \sum_{\substack{j,k=1\\ j \neq k}}^n \xi_j \xi_k B_{j,k} \right) \right]
& \leq 4 ~\E_{\Xi |F} \left[  \sup_{B \in {\cal B}} \E_{\Delta} h \left(  \sum_{\substack{j,k=1\\ j \neq k}}^n \delta_j (1-\delta_k) \xi_j \xi_k B_{j,k} \right) \right]~,
\end{align*}
where we have taken conditional expectations $\Xi | F$.

Consider a fixed realization $\Delta_r = \{\delta_{1,r},\ldots,\delta_{n,r}\}$ of $\Delta$, and consider the subset $I = \{j \in [n] | \delta_{j,r} = 1\}$. Lets $I^c$ be the complement set. Then,
\beq
4 \left[ \sum_{\substack{j,k=1\\ j \neq k}}^n \delta_{j,r} (1-\delta_{k,r}) \xi_j \xi_k B_{j,k} \right]
= 4 \left[ \sum_{\substack{(j,k) \in I \times I^c \\ j \neq k}} \xi_j \xi_k B_{j,k} \right]~.
\eeq
Since $\Xi' = \{ \xi'_i \}$ is a decoupled tangent sequence to $\Xi = \{ \xi_i \}$, by Proposition~\ref{prop:dist1}, we have 
%for any $(j,k)$ with $j < k$, $(\xi_j,\xi_k)$ and $(\xi_j,\xi'_k)$ are identically distributed.
%As a result,
\begin{align}
\E_{\Xi|F} \left[ \sup_{B \in {\cal B}} h \left( \sum_{\substack{j,k=1\\ j\neq k}}^n \xi_j \xi_k B_{j,k} \right) \right]
& \leq 4 ~ \E_{\Xi|F} \left[  \sup_{B \in {\cal B}} \E_{\Delta} h \left(  \sum_{\substack{j,k=1\\ j \neq k}}^n \delta_i (1-\delta_j) \xi_j \xi_k B_{j,k} \right) \right] \nonumber \\
& = 4 ~ \E_{\Xi|F} \left[  \sup_{B \in {\cal B}} \E_{\Delta} h \left(  \sum_{\substack{(j,k) \in I \times I^c \\ j \neq k}} \xi_j \xi_k B_{j,k} \right) \right] \\
& \stackrel{(a)}{=} 4 ~\E_{\Xi, \Xi'|F} \left[  \sup_{B \in {\cal B}} \E_{\Delta} h \left( \sum_{\substack{(j,k) \in I \times I^c \\ j \neq k}} \xi_j \xi'_k B_{j,k} \right) \right] ~.
\label{eq:dec1}
\end{align}
where (a) follows from the fact that if two random variables are identically distributed, expectations of the same function applied to them will be the same. Note that the relevant matrix $\tilde{B}$ for Proposition~\ref{prop:dist1} here is $\tilde{B}_{j,k} = B_{j,k}$ for $(j,k) \in I  \times I^c, j \neq k$ and $\tilde{B}_{j,k} = 0$ otherwise. 

Let 
\beq
Y(\Delta) \triangleq 4 \sum_{\substack{j,k=1\\j \neq k\\(j,k) \in I \times I^c}}^n \xi_j \xi'_k B_{j,k}~, \quad
Z(\Delta) \triangleq 4 \sum_{\substack{j,k=1\\j \neq k\\(j,k) \not \in I \times I^c}}^n \xi_j \xi'_k B_{j,k}~, \quad
W \triangleq 4 \sum_{j=1}^n \xi_j \xi'_j B_{j,j}~.
\eeq
By construction, for every realization $\Delta_r$, we have
\beq
Y(\Delta_r) + Z(\Delta_r) + W = 4 \left[ \sum_{j,k=1}^n \xi_j \xi'_k B_{j,k} \right]~.
\eeq
Now, by linearly of expectation, we have
\begin{align*}
\E_{\Xi,\Xi'|F}[Z + W] & = 4 \sum_{\substack{j,k=1\\j \neq k\\(j,k) \not \in I \times I^c}}^n \E_{\xi_j,\xi'_k|F} [\xi_j \xi'_k] B_{j,k} + 4 \sum_{j=1}^n \E_{\xi_j,\xi'_j|F}[\xi_j \xi'_j] B_{j,j} \\
& = 4 \sum_{\substack{j,k=1\\j \neq k\\(j,k) \not \in I \times I^c}}^n \E_{\xi_j|F} [\xi_j] \E_{\xi'_k|F}[ \xi'_k] B_{j,k} + 4 \sum_{j=1}^n \E_{\xi_j|F}[\xi_j] \E_{\xi'_j|F} [\xi'_j] B_{j,j} \\
& = 0~.
\end{align*}
\begin{comment}
We focus on one term $\E_{\xi_j,\xi'_k|F} [\xi_j \xi'_k]$. For  $j < k$, we have
\begin{align*}
\E_{\xi_j,\xi'_k|F} [\xi_j \xi'_k] & = \E_{\xi_{1:(k-1)}}\left[ E_{\xi_j,\xi'_k} \left[\xi_j \xi'_k | \xi_{1:(k-1)}\right]\right] 
= E_{\xi_{1:(k-1)}}\left[ \xi_j  E_{\xi'_k}\left [\xi'_k | \xi_{1:(k-1)}\right]\right] = 0~,
\end{align*}
since $\xi'_k|\xi_{1:(k-1)}$ is a martingale difference sequence, which has zero mean. The argument for $j > k$ is similar by interchanging $\Xi$ and $\Xi'$. Recall that $\Xi,\Xi'$ are decoupled tangent sequences, and, following Proposition~\ref{prop:dec4}, let $\cG = \sigma(\{\xi_j\})$ be the master $\sigma$-field with respect to which $\{\xi'_j\}$ are conditionally independent. Then, we have
\begin{align*}
E_{\xi_j,\xi'_j} [\xi_j \xi'_j] & = E_{\cG}\left[ E_{\xi_j,\xi'_j} \left[\xi_j \xi'_j | \cG \right]\right] 
%\\
%& 
\stackrel{(a)}{=} E_{\cG}\left[ \xi_j  E_{\xi'_j}\left [\xi'_j | \cG \right]\right] 
%\\
%& 
\stackrel{(b)}{=} E_{\cG}\left[ \xi_j  E_{\xi'_j}\left [\xi'_j | \cF_{j-1} \right]\right] 
%\\
%& 
\stackrel{(c)}{=} 0~,
\end{align*}
where (a) follows since $\xi_j$ if $\cG$-measurable, (b) follows since $P(\xi'_j|\cG) = p(\xi'_j|\cF_{j-1})$ from Definition~\ref{defn:dec2}, and (c) follows since $\xi'_j|\cF_{j-1}$ is a MDS.
As a result, it follows that 
\beq
E_{\Xi,\Xi'}[Z+W] = 0~.
\eeq
\end{comment}

Now, for any convex function $h$, by Jensen's inequality, we have
\begin{equation*}
\E_{\Xi,\Xi'|F} [h(Y)] = \E_{\Xi,\Xi'|F} [h(Y + \E_{\Xi,\Xi'|F}[Z+W])] \leq \E_{\Xi,\Xi'|F} [h(Y + Z + W)]~.
\end{equation*}
%\ab{An odd application of Jensen's -- does this make sense?}
%\eeq
%Further, for every realization $\Delta_r$, we have $Z(\Delta_r) = 0$, so $P(Z(\Delta)=0) = 1$ and $E_{\Delta}[Z(\Delta)] = 0$. Note that $Z(\Delta)$ is independent of $Y(\Delta)$. Again, by Jensen's inequality,
%\begin{align*}
%E_{\Delta} F \left( Y(\Delta) \right| & =  E_{\Delta} \left| Y(\Delta) + E_{\Delta}[Z(\Delta)] \right| \\
%& \leq E_{\Delta} \left| Y(\Delta) + Z(\Delta) \right| \\
%\Rightarrow \qquad
%E_{\Delta} \left| 4 \sum_{(j,k) \in I \times I^c} \xi_j \xi'_k B_{j,k} \right|
%& \leq E_{\Delta} \left| 4 \sum_{\substack{(j,k)=1\\j \neq k}}^n \xi_j \xi'_k B_{j,k} \right|
% = \left| 4 \sum_{\substack{(j,k)=1\\j \neq k}}^n \xi_j \xi'_k B_{j,k} \right|
%\end{align*}
Then, from \eqref{eq:dec1}, we have 
\begin{align*}
E_{\Xi|F} \left[ \sup_{B \in {\cal B}} h \left( \sum_{\substack{j,k=1\\ j\neq k}}^n \xi_j \xi_k B_{j,k} \right) \right]
& \leq 4 E_{\Xi, \Xi'|F} \left[  \sup_{B \in {\cal B}} E_{\Delta} h \left(  \sum_{\substack{(j,k) \in I \times I^c \\ j \neq k}} \xi_j \xi'_k B_{j,k} \right) \right] \\
& \leq 4 E_{\Xi, \Xi'|F} \left[  \sup_{B \in {\cal B}} E_{\Delta} h \left(  \sum_{\substack{j,k=1}}^n \xi_j \xi'_k B_{j,k} \right) \right] \\
& = 4 E_{\Xi, \Xi'|F} \left[  \sup_{B \in {\cal B}} h \left(  \sum_{\substack{j,k=1}}^n \xi_j \xi'_k B_{j,k} \right) \right]~.
\end{align*}
Taking expectations w.r.t.~$F$ on both sides completes the proof. \qed

\subsection{Classical Construction of Decoupled Tangent Sequences}
\label{ssec:dtsold}
Our use of the phrase `decoupled tangent sequence' is inspired by related developments in the context of decoupling for martingales. We briefly revisit this usage and a useful result in this context. Our exposition is based on the classical text on decoupling~\cite{pegi99}, especially Chapter 6. 
%The following definitions are from \cite[Chapter 6]{pegi99}.
\begin{defn}
Let $\{e_i\}$ and $\{d_i\}$ be two sequences of random variables
adapted to the $\sigma$-fields $\{ \cF_i \}$. Then $\{e_i\}$ and $\{d_i\}$ are tangent with respect to $\{ \cF_i \}$ if, for all $i$,
\beq
p(d_i| \cF_{i-1}) = p(e_i| \cF_{i-1})~,
\eeq
where $p(d_i|\cF_{i-1})$ denotes the conditional probability of $d_i$ given $\cF_{i-1}$.
\label{defn:dec1}
\end{defn}

\begin{defn}
A sequence $\{e_i\}$ of random variables adapted to an increasing
sequence of $\sigma$-fields $\cF_i$ contained in $\cF$ is said to satisfy the CI condition (conditional
independence) if there exists a $\sigma$-algebra $\cG$, contained in $\cF$ such that $\{ e_i \}$  is
conditionally independent given $\cG$, and $p(e_i|\cF_{i-1}) = p(e_i|\cG)$.
\label{defn:dec2}
\end{defn}

\begin{defn}
A sequence $\{ e_i \}$ which satisfies the CI condition and which is
also tangent to $\{ d_i \}$ is said to be a decoupled tangent sequence to $\{d_i\}$.
\label{defn:dec3}
\end{defn} 

The following result is from \cite[Proposition 6.1.5]{pegi99}.
\begin{prop}
For any sequence of random variables $\{d_i\}$ adapted to an
increasing sequence $\cF_i$ of a $\sigma$-algebras, there always exists a decoupled sequence
$\{e_i\}$ (on a possibly enlarged probability space) which is tangent to the original
sequence and in addition conditionally independent given a master $\sigma$-field $\cG$.
Frequently $\cG = \sigma(\{d_i\})$.
\label{prop:dec4}
\end{prop}

% \section{A Special Case: Rademacher MDS}
% \label{sec:rip-rademacher}
% \input{sec/rip-rademacher}

\section{Bounds for Stochastic Processes}
\label{sec:rip-subg}

\subsection{Overall Analysis}
\label{ssec:subg-main}

For a stochastic process $\bxi = \{\xi_j\}$ adapted to $\{F_i\}$ satisfying (SP-1) and (SP-2), let
\begin{align}
    C_{\cA}(\bxi) & \triangleq ~\sup_{A \in \cA} \left| \|A\bxi\|_2^2 - E\|A\bxi\|_2^2 \right| \\
    B_{\cA}(\bxi) & \triangleq ~\sup_{A \in \cA} \left| \sum_{\substack{j,k=1\\j \neq k}}^n \xi_j \xi_k \langle A_j, A_k \rangle \right| \\
    D_{\cA}(\bxi) & \triangleq ~\sup_{A \in \cA} \left| \sum_{j=1}^n (|\xi_j|^2 - E|\xi_j|^2) \| A_j \|_2^2 \right| 
    % \\    
    % N_{\cA}(\bxi) & \triangleq ~\sup_{A \in \cA} \| A \bxi \|_2~.
\end{align}
First, note that the contributions from the off-diagonal terms of $E\|A\bxi\|_2^2$ is 0:
\begin{prop}
For $j \neq k$, $\E_{\xi_j,\xi_k}[\xi_j \xi_k] = 0$.
\end{prop}
\proof Since $\{\xi_i\}$ is a stochastic process adapted to $\{F_i\}$ satisfying (SP-1) and (SP-2), with $F=F_{1:n}$ we have
\begin{align*}
\E_{\xi_j,\xi_k}[\xi_j \xi_k] & = \E_F[\E_{\xi_j,\xi_k|F}[\xi_j \xi_k] = \E_F[\E_{\xi_j|F}[\xi_j] \E_{\xi_j|F}[\xi_k]]\\
& = 0~,
\end{align*}
since $\xi_j \perp \xi_k | F$ by (SP-2) and $\E_{\xi_j|F}[\xi_j] = 0 = \E_{\xi_k|F}[\xi_k]$ by (SP-1). \qed

% For  $j < k$, we have
% \begin{align*}
% E_{\xi_j,\xi_k} [\xi_j \xi_k] & = E_{\cF_{k-1}}\left[ E_{\xi_j,\xi_k} \left[\xi_j \xi_k | \cF_{k-1}\right]\right] 
% = E_{\cF_{k-1}}\left[ \xi_j  E_{\xi_k}\left [\xi_k | \cF_{k-1}\right]\right] = 0~,
% \end{align*}
% since $\xi_k|\cF_{k-1}$ is a martingale difference sequence, which has zero mean. The proof for $j > k$ is similar by switching the roles of $j$ and $k$.  \qed

As a result, we have
\begin{align*}
    C_{\cA}(\bxi) & = ~ \sup_{A \in \cA} \left| \|A\bxi\|_2^2 - E\|A\bxi\|_2^2 \right| \\
    & =~ \sup_{A \in \cA} \left| \sum_{\substack{j,k=1\\j \neq k}}^n \xi_j \xi_k \langle A_j, A_k \rangle + \sum_{j=1}^n (|\xi_j|^2 - E|\xi_j|^2) \| A_j \|_2^2 \right| \\
    & \leq~ \sup_{A \in \cA} \left| \sum_{\substack{j,k=1\\j \neq k}}^n \xi_j \xi_k \langle A_j, A_k \rangle \right| + \sup_{A \in \cA} \left| \sum_{j=1}^n (|\xi_j|^2 - E|\xi_j|^2) \| A_j \|_2^2 \right| \\
    & =~ B_{\cA}(\bxi) +  D_{\cA}(\bxi) 
\end{align*}    
Hence,
\begin{align}
    \| C_{\cA}(\bxi) \|_{L_p} & \leq \| B_{\cA}(\bxi)\|_{L_p}  +  \| D_{\cA}(\bxi) \|_{L_p}~.
\end{align}

We bound $\| B_{\cA}(\bxi)\|_{L_p}$ in Section~\ref{ssec:offd} (Theorem~\ref{theo:offd}) and bound $\| D_{\cA}(\bxi) \|_{L_p}$ in Section~\ref{ssec:diag} (Theorem~\ref{theo:subg-diag}) to get a bound on $\| C_{\cA}(\bxi)\|_{L_p}$  of the form
\begin{equation}
    \| C_{\cA}(\bxi)\|_p \leq a + \sqrt{p} \cdot b + p \cdot c~, \quad \forall p \geq 1~,
    \label{eq:lpp1}
\end{equation}
where $a,b,c$ are constants independent of $p$. Note that these bounds imply, for all $u$
\begin{equation}
    P(\left| C_{\cA}(\bxi) \right|  \geq a + b \cdot \sqrt{u} + c \cdot u ) \leq e^{-u}~,
    \label{eq:lpp2}
\end{equation}
or, equivalently
\begin{equation}
    P(\left| C_{\cA}(\bxi) \right|  \geq a + u ) \leq \exp\left\{-\min\left(\frac{u^2}{4b^2}, \frac{u}{2c}\right)\right\}~,
    \label{eq:lpp3}
\end{equation}
which yields the main result. In the sequel, to avoid clutter, we mostly avoid all absolute constants and constants which depend on $L$ for $L$-sub-Gaussian random variables, i.e., we set them to 1, so the key dependencies are clear. We are inspired by similar choices in the related literature~\cite{tala14,krmr14}, where $c$ is used to denote constants which may keep changing from one line to the next.

%\ab{The last statement, formally stated, will be the main theorem.}

%\ab{Applications of the result to say restricted isometry (RIP) or johnson-lindenstrauss (J-L) can now be obtained following standard approaches, e.g., see Jelani Nelson's notes, [KMR14], etc.}

%\subsection{The Scaling Term}

\subsection{The Off-Diagonal Terms}
\label{ssec:offd}
The main result for the off-diagonal term is the following:
\begin{theo}
    Let $\bxi$ be a stochastic process adapted to $F = \{F_i\}$ satisfying (SP-1) and (SP-2). Then, for all $p \geq 1$, we have
\begin{align*}
    \| B_{\cA}(\bxi)\|_p & \leq \gamma_2(\cA,\| \cdot \|_{2 \rightarrow 2}) \cdot \bigg( \gamma_2(\cA,\| \cdot \|_{2 \rightarrow 2}) + d_F(\cA) \bigg) \\ 
    & \phantom{....} + \sqrt{p} \cdot d_{2 \rightarrow 2}(\cA) \cdot \bigg( \gamma_2(\cA,\| \cdot \|_{2 \rightarrow 2}) + d_F(\cA) \bigg) + p \cdot d_{2 \rightarrow 2}^2 (\cA)~.
\end{align*}
\label{theo:offd}
\end{theo}
Note that from the main decoupling result in Theorem~\ref{th:mds-dec}, choosing $h(x) = |x|^p, p \geq 1$ as the convex function, applying the decoupling inequality, and taking $p$-th root on both sides, we have
\begin{align}
\| B_{\cA}(\bxi) \|_{L_p} & \leq  \left\| \sup_{A \in \cA} \left| \sum_{j,k=1}^n \xi_j \xi'_k \langle A_j, A_k \rangle \right| \right\|_{L_p} = \left\| \sup_{A \in \cA} \left| \langle A \bxi, A \bxi' \rangle \right| \right\|_{L_p}~.
\label{eq:bp1}
\end{align}
Hence our analysis will focus on bounding \eqref{eq:bp1}, the $L_p$-norm of the decoupled quadratic form.
We start with the following result: %for the proof of Theorem~\ref{theo:offd}.
\begin{lemm}
Let $\bxi = \{\xi_i\}$ be a stochastic process adapted to $F$ satisfying (SP-1) and (SP-2), and $\bxi'$ be a decoupled tangent sequence to $\bxi$. Then, for every $p \geq 1$,
\begin{equation}
    \left\| \sup_{A \in \cA} \langle A \bxi, A \bxi' \rangle \right\|_{L_p}
    \leq \gamma_2(\cA,\| \cdot \|_{2 \rightarrow 2}) \cdot \| N_{\cal A}(\bxi) \|_{L_p} + \sup_{A \in \cA} \| \langle A \bxi, A \bxi' \rangle \|_{L_p}~,
\end{equation}
where $N_{\cA}(\bxi) = \sup_{A \in \cA} \| A \bxi \|_2$.
\label{lemm:offd-lp}
\end{lemm}

{\em Proof of Lemma~\ref{lemm:offd-lp}:} Without loss of generality, assume $\cA$ is finite~\cite{tala14}. Consider the random variable of interest: 
\begin{align*}
    \Gamma =  \sup_{A \in \cA} \left| \langle A \xi, A \bxi' \rangle \right|~.
\end{align*}
Let $\{T_r\}_{r=0}^{\infty}$ be an admissible sequence for $\cA$ for which the minimum in the definition
of $\gamma_2(\cA, \| \cdot \|_{2 \rightarrow 2})$ is attained. Let 
\begin{align*}
    \pi_r A = d_{2 \rightarrow 2}(A,T_r) = \underset{B \in T_r}{\argmin} ~\| B - A\|_{2 \rightarrow 2} \qquad \text{and} \qquad \Delta_r A = \pi_r A - \pi_{r-1} A~.
\end{align*}
For any given $p \geq 1$, let $\ell$ be the largest integer for which $2^{\ell} \leq 2 p$.
Then, by a direct computation based on a telescoping sum and application of triangle inequality, we have 
\beq
\left| \langle A \bxi, A \bxi'\rangle - \langle (\pi_{\ell} A) \bxi, (\pi_{\ell} A) \bxi' \rangle \right|
\leq \underbrace{\left| \sum_{r=\ell}^{\infty} \langle (\Delta_{r+1} A) \bxi, (\pi_{r+1} A) \bxi' \rangle \right|}_{S_1}  
+ \underbrace{\left| \sum_{r=\ell}^{\infty} \langle (\pi_{r} A) \xi, (\Delta_{r+1} A) \xi' \rangle \right|}_{S_2}~. 
\label{eq:l3tri}
\eeq
We focus on $S_1$ noting that the analysis for $S_2$ is similar. Let
\begin{equation*}
    X_r(A) = \langle (\Delta_{r+1} A) \bxi, (\pi_{r+1} A) \bxi' \rangle~. 
\end{equation*}
Conditioning $X_r(A)$ on $\bxi'$ and $F$, we note 
\begin{equation*}
 X_r(A)|F =   \langle (\Delta_{r+1} A) \bxi, (\pi_{r+1} A) \bxi' \rangle|F = \langle  \bxi, (\Delta_{r+1} A)^T (\pi_{r+1} A) \bxi' \rangle|F
\end{equation*}
a weighted sum of a sub-Gaussian martingale difference sequence. Then, a direct application of the Azuma-Hoeffding bound~\cite{bolm13} gives
\begin{align*}
    P\left( |X_r(A) | > u \| (\Delta_{r+1} A)^T (\pi_{r+1} A) \bxi' \|_2 ~\bigg| ~\bxi',F \right) \leq 2 \exp(-u^2/2)~.
\end{align*}
Using $u = t 2^{r/2}$, we get
\begin{align*}
P\left( |X_r(A) | > t 2^{r/2} \| (\Delta_{r+1} A)^T (\pi_{r+1} A) \bxi' \|_2 ~\bigg|~ \bxi', F \right) \leq 2 \exp(-t^2 2^r/2)~.
\label{eq:azho}
\end{align*}
Since
\begin{align*}
\left| (\Delta_{r+1} A)^T (\pi_{r+1} A) \bxi' \right| 
\leq \| \Delta_{r+1} A \|_{2 \rightarrow 2} \sup_{A \in \cA} \| A \bxi' \|_2~.
\end{align*}
we have 
\begin{align*}
P\left( |X_r(A) | > t 2^{r/2} \| \Delta_{r+1} A \|_{2 \rightarrow 2} \sup_{A \in \cA} \| A \bxi' \|_2 ~\bigg|~ \bxi', F \right) \leq 2 \exp(-t^2 2^r/2)~.
\end{align*}
Now, since $|\{\pi_r A : A \in \cA \}| = |T_r| \leq 2^{2^r}$, by union bound, we get 
 \begin{align*}
P\bigg( \sup_{A \in \cA}  &  ~\sum_{r=\ell}^{\infty} |X_r(A)| > t \left( \sup_{A \in \cA} \sum_{r=\ell}^{\infty} 2^{r/2} \| \Delta_{r+1} A \|_{2 \rightarrow 2} \right) \cdot \sup_{A \in \cA} \| A \bxi' \|_2  ~\bigg|~ \bxi', F \bigg) \\  
 & \leq 2 \sum_{r=\ell}^{\infty} |T_{r}| \cdot |T_{r+1}| \cdot \exp(-t^2 2^r/2) \\
 & \leq 2 \sum_{r=\ell}^{\infty} 2^{2^{r+2}} \cdot \exp (-t^2 2^r/2) \\
 & \leq 2 \exp(-2^{\ell} t^2)~,
% \label{eq:bnd42}
\end{align*}
for all $t \geq t_0$, a constant. Noting that
\begin{align*}
    \sup_{A \in \cA} \sum_{r=\ell}^{\infty} 2^{r/2} \| \Delta_{r+1} A \|_{2 \rightarrow 2} & = \gamma_2(\cA, \| \cdot \|_{2 \rightarrow 2} ) \\
    \sup_{A \in \cA} \| A \bxi' \|_2 & = N_{\cA}(\bxi')~,
\end{align*}
we have 
\begin{align*}
P\bigg( \sup_{A \in \cA} ~\sum_{r=\ell}^{\infty} |X_r(A)| > t \gamma_2(\cA,  \|\cdot \|_{2 \rightarrow 2}) N_{\cA}(\bxi')  ~\bigg|~ \bxi',F \bigg) 
\leq 2 \exp(-p t^2)~,
\end{align*}
since $p \leq 2^{\ell}$ by construction. In other words, with $V(\bxi') = \gamma_2(\cA,  \|\cdot \|_{2 \rightarrow 2}) N_{\cA}(\bxi')$, for $t \geq t_0$ we have
\begin{align*}
    P\left( S_1 \geq t V(\bxi') ~\bigg|~ \bxi',F \right) \leq 2 \exp(- p t^2)~.
\end{align*}
Note that
\begin{align*}
    \| S_1 \|_{L_p}^p & = \E_{\bxi,\bxi'} S_1^p = \E_F \left[ \E_{\bxi,\bxi'|F} \left[S_1^p \right] \right] = E_{\bxi',F} \int_0^{\infty} p t^{p-1} P(S_1 > t ~\big|~ \bxi',F) dt~,
\end{align*}
and
\begin{align*}
    \int_0^{\infty} p t^{p-1} P(S_1 > t ~\big|~ \bxi', F) dt & \leq c^p V(\bxi')^p + \int_{cV(\bxi')}^{\infty} p t^{p-1} P(S_1 > t ~\big|~ \bxi',F) dt \\
    & \leq c^p V(\bxi')^p + V(\bxi')^p \int_{c}^{\infty} p \tau^{p-1} P( S_1 > \tau V(\bxi') | \bxi',F) d\tau \\
    & \leq c_1^p V(\bxi')^p~,
\end{align*}
where $c \geq t_0,c_1$ are suitable constants with depend on $L$. As a result, $\| S_1 \|_{L_p} \leq c_1 V(\bxi') = c_1 V(\bxi)$.
The bound on $\| S_2 \|_{L_p}$ is the same, and can be derived similarly. As a result
\begin{equation}
    \| S_1 + S_2 \|_{L_p}  \leq c_2 \gamma_2(\cA,  \|\cdot \|_{2 \rightarrow 2}) \| N_{\cA}(\bxi) \|_{L_p}
    \label{eq:s1s2}
\end{equation}
Further, since $| \{ \pi_{\ell} A : A \in \cA \} | \leq 2^{2^{\ell}} \leq \exp(2p)$, we have 
\begin{align*}
    E \sup_{A \in \cA} | \langle (\pi_{\ell} A) \bxi, (\pi_{\ell} A) \bxi' \rangle |^p & \leq \sum_{A \in T_{\ell}} E| \langle A\bxi, A \bxi'|^p \leq 2^{2p} \sup_{A \in \cA} E | \langle A \bxi, A \bxi' \rangle |^p~,
\end{align*}
so that
\begin{equation}
    \left\| \sup_{A \in \cA} | \langle (\pi_{\ell} A) \bxi, (\pi_{\ell} A) \bxi' \rangle \right\|_{L_p} \leq 4 \| \sup_{A \in \cA} E | \langle A \bxi, A \bxi' \rangle \|_{L_p}~.
    \label{eq:s0}
\end{equation}
Combining \eqref{eq:l3tri}, \eqref{eq:s1s2}, and \eqref{eq:s0} using triangle inequality completes the proof. \qed

%\proof By generic chaining and Azuma-Hoeffding. \qed

For the first term in Lemma~\ref{lemm:offd-lp}, we have the following bound:
\begin{lemm}
Let $\bxi$ be a stochastic process adapted to $F$ satisfying (SP-1) and (SP-2) and let $N_{\cA}(\bxi) = \sup_{A \in \cA} \| A \bxi \|_2$. Then
\begin{equation}
\| N_{\cA}(\bxi) \|_{L_p} \leq \gamma_2(\cA, \| \cdot \|_{2 \rightarrow 2}) + d_F(\cA) + \sqrt{p} d_{2 \rightarrow 2}(\cA)~.
\end{equation}
\label{lemm:scaling}
\end{lemm}
\proof Consider the set $S = \{ A^T x : x \in B_2^n, A \in \cA \}$. Since $\bxi$ satisfies (SP-1), it is sub-Gaussian conditioned on $F$ and we have
\begin{align*}
    \| N_{\cA}(\bxi) \|_{L_p} & = \left( \E_{\bxi} \left[ \sup_{A \in \cA, x \in B_2^n} | \langle A\bxi, x \rangle |^p \right] \right)^{1/p} = \left( \E_{\bxi} \left[\sup_{u \in S} | \langle \bxi, \u \rangle |^p \right] \right)^{1/p} \\
    & \stackrel{(a)}{\leq} \E_{\bgg} \left[ \sup_{u \in S} |\langle u, \bgg \rangle | \right] + \sup_{u \in S} ~(\E_{\bxi} | \langle \bxi, u \rangle |^p )^{1/p} \\
    & = \E_{\bgg} \left[ \sup_{A \in \cA, x \in B_2^n} | \langle A\bgg, x \rangle | \right] + \sup_{u \in S} ~(\E_F \E_{\bxi|F} | \langle \bxi, u \rangle |^p )^{1/p} \\
    & = \E_{\bgg} \left[ \sup_{A \in \cA} N_{\cal A}(\bgg) \right] + \sqrt{p} \sup_{A \in \cA, x \in B_2^n} \| A^T x \|_2 \\
    & \stackrel{(b)}{\leq} \gamma_2(\cA,\| \cdot \|_{2 \rightarrow 2}) + d_F(\cA) + \sqrt{p} \cdot d_{2 \rightarrow 2}(\cA)~,
\end{align*}
where (a) follows from Lemma~\ref{lemm:lpnorm} and (b) follows from \cite[Lemma 3.7]{krmr14}. \qed

For the second term, we have the following bound:
\begin{lemm}
    Let $\bxi$ be a stochastic process adapted to $F$ satisfying (SP-1) and (SP-2), and let $\bxi'$ be a decoupled tangent sequence. Then, for every $p \geq 1$,
    \begin{equation}
        \sup_{A \in \cA} \| \langle A \bxi, A \bxi' \rangle \|_{L_p} \leq 
        \sqrt{p} \cdot d_F(\cA) \cdot d_{2 \rightarrow 2}(\cA) + p \cdot d_{2 \rightarrow 2}^2(\cA)~.
    \end{equation}
\label{lemm:offd-lp-2nd}
\end{lemm}
Proof of Lemma~\ref{lemm:offd-lp-2nd} needs the following result:
\begin{lemm}
    Let $\x_1,\ldots,\x_n \in \R^d$ and $T  \subset \R^d$. Let $\bxi = \{ \xi_j\}$ be a stochastic process adapted to $F=\{F_i\}$ satisfying (SP-1) and (SP-2), and let $\y = \sum_{j=1}^n \xi_j \x_j$. Then, for every $p \geq 1$,
    \begin{align}
        \left( \E_{\bxi} \left[ \sup_{t \in T} | \langle t, \y \rangle |^p \right] \right)^{1/p} & \leq c_2 \left( \E_{\bgg} \left[ \sup_{t \in T} | \langle t, \bgg \rangle | \right] + \sup_{t \in T} \left( \E_{\bxi} \left[ | \langle t, \y \rangle|^p \right] \right)^{1/p}  \right)~ 
    \end{align}
    where $c_2$ is a constant which depends on $L$ and $\bgg = \sum_{j=1}^n g_j \x_j$ where $g_i \sim N(0,1)$ are independent.
\label{lemm:lpnorm}
\end{lemm}

We need the following basic property of sub-Gaussian random variables~\cite{vers14} to prove Lemma~\ref{lemm:lpnorm}.

\begin{prop}
If $X$ is a $L$-sub-Gaussian random variable, then for some suitable constant $c_0$ which depends on $L$, we have
\begin{equation}
    P(|X| > t L) \leq 2 \exp(-t^2)~, ~~\forall t \geq 0 \qquad \Leftrightarrow \qquad  (E|X|^p)^{1/p} \leq c_0 \sqrt{p} L~, ~~\forall p~. 
\end{equation}
\label{prop:subg}
\end{prop}

{\em Proof of Lemma~\ref{lemm:lpnorm}.} We assume $T$ is finite without loss of generality~\cite{tala14}. Let $\{T_r\}$ be an optimal admissible sequence of $T$. For any $t \in T$, let $\pi_r(t) = \argmin_{t_r \in T_r} \| t - t_r\|_2$. For any given $p$ determining the $p$-norm, choose $\ell$ such that $2^{\ell-1} \leq 2p \leq 2^{\ell}$, so that $2^{\ell} / p \leq 4$. Then, by triangle inequality, we have
\begin{equation}
    \sup_{t \in T}|\langle t,\y \rangle| \leq \sup_{t \in T} | \langle \pi_{\ell}(t), \y \rangle | + \sup_{t\in T} \sum_{r=\ell}^{\infty} | \langle \pi_{r+1}(t) - \pi_r(t), \y \rangle |~.
\end{equation}
For the first term, note that
\begin{align*}
    \left( \E_{\bxi} \left[ \sup_{t \in T} | \langle \pi_{\ell}(t), \y \rangle |^p \right] \right)^{1/p} 
    & \leq \left( \E_{\bxi} \left[ \sum_{t \in T_{\ell}} | \langle t, \y \rangle |^p \right] \right)^{1/p} \\
%    & = \left( \E_F \left[ \E_{\bxi|F} \left[ \sum_{t \in T_{\ell}} | \langle t, \y \rangle |^p \right] \right] \right)^{1/p} \\
    & \leq (|T_{\ell}|)^{1/p} \sup_{t \in T_{\ell}} \left(\E_{\bxi}  | \langle t, \y \rangle |^p  \right)^{1/p} \\
    & \leq (2^{2^{\ell}})^{1/p} \sup_{t \in T} (\E_{\bxi} |\langle t,\y \rangle|^p)^{1/p} \\
    & \leq 16 \sup_{t \in T} ( \E_{\bxi} |\langle t,\y \rangle|^p)^{1/p}~.
\end{align*}
For the second term, since $\{\xi_j\}$ is a stochastic process satisfying (SP-1) and (SP-2), 
for any fixed realization $f_{1:n}$ of $F$, we have 
\begin{align*}
    \P_{\bxi|f_{1:n}} & \left( \sup_{t\in T} \sum_{r=\ell}^{\infty} | \langle \pi_{r+1}(t) - \pi_r(t), \y \rangle | \geq u L \sum_{r=\ell}^{\infty} 2^{r/2} \| (\langle \pi_{r+1}(t) - \pi_r(t), \x_j \rangle)_{j=1}^n \|_2 \right) \\
    & \leq \sum_{r=\ell}^{\infty} \sum_{t \in T_{r+1}} \sum_{t' \in T_r} \P_{\bxi|f_{1:n}} \left( \left| \sum_{j=1}^n \xi_j \langle t - t', \x_j \rangle \right| \geq u L 2^{r/2} \| \langle t - t', \x_j \rangle_{j=1}^n \|_2 \right) \\
    & \stackrel{(a)}{\leq} \sum_{r=\ell}^{\infty} 2^{2^{r+1}} \cdot 2^{2^r} \cdot \exp (-2^r u^2/2) \leq 2 \exp(- 2^{\ell} u^2/4) \\
    & \leq 2 \exp(-pu^2/2)~,
\end{align*}
for $u > c$, a constant (see Remark on generic chaining union bound in the sequel), where (a) follows from Hoeffding inequality. Since the result holds for any realization $f_{1:n}$, taking expectation w.r.t.~$F$ to remove the conditioning, we have 
\begin{align*}
  \P_{\bxi} & \left( \sup_{t\in T} \sum_{r=\ell}^{\infty} | \langle \pi_{r+1}(t) - \pi_r(t), \y \rangle | \geq u L \sum_{r=\ell}^{\infty} 2^{r/2} \| (\langle \pi_{r+1}(t) - \pi_r(t), \x_j \rangle)_{j=1}^n \|_2 \right) \\
    & \leq 2 \exp(-pu^2/2)~.
\end{align*}
Then, from Proposition~\ref{prop:subg}, we have
\begin{align*}
    \left( \E_{\bxi} \sup_{t\in T} \sum_{r=\ell}^{\infty} | \langle \pi_{r+1}(t) - \pi_r(t), \y \rangle |^p \right)^{1/p} 
    & \leq L \sum_{r=\ell}^{\infty} 2^{r/2} \| (\langle \pi_{r+1}(t) - \pi_r(t), \x_j \rangle)_{j=1}^n \|_2 \\
    & \leq L \gamma_2(T', \| \cdot \|_2 )~,
\end{align*}
where $T' = \{ (\langle t, \x_j \rangle)_{j=1}^n | t \in T \}$. Then, by the majorizing measures theorem~\cite{tala14,tala05}, we have
\begin{equation*}
    \gamma_2(T',\| \cdot \|_2 ) \leq E \sup_{t' \in T'} \left| \langle t', \bgg \rangle \right| 
    = E \sup_{t \in T} \left| \sum_{j=1}^n \langle t, \x_j \rangle g_j \right| = E \sup_{t \in T} |  \langle t, \bgg \rangle |~,
\end{equation*}
where $\bgg = \sum_{j=1}^n g_j \x_j$. That completes the proof. \qed

Before proceeding further, we show the details of how the union bound works out in generic chaining~\cite{tala14}. We use variants of such union bound analysis several times in our proofs, and this is the only place we show the details. Such analysis is considered standard in the context of generic chain, but as a tool generic chaining is not as widely used.
%which are in fact quite straightforward.

{\bf Remark: Union bound in generic chaining.} After applying union bound in a generic chaining based analysis, we get a (infinite) sum of the following form:
\begin{align*}
    \sum_{r=\ell}^{\infty} 2^{2^{r+1}} \cdot 2^{2^r} \cdot \exp (-2^r u^2/2) 
    & = \sum_{r=\ell}^{\infty} 2^{3\cdot 2^{r}} \cdot \exp( -  2 \cdot 2^{r} u^2/4) \\
    & = \exp(-2^{\ell} u^2/4) \sum_{r=\ell}^{\infty} \exp^{(3 \log 2) \cdot 2^r} \cdot \exp( - 2 \cdot (2^r - 2^{\ell}) u^2/4)~. 
\end{align*}
Focusing on the exponent, note that
\begin{align*}
    (3 \log 2) \cdot 2^r - 2 \cdot 2^r u^2/4 +  \cdot 2^{\ell} u^2 /4  & < - (r-\ell) \\ 
    \Rightarrow \quad - (2^{r+1} - 2^{\ell}) u^2 /2 & < - (r - \ell) - (3 \log 2) \cdot 2^r \\
    \Rightarrow \quad  (2^{r+1} - 2^{\ell}) u^2 /2 & > (r - \ell) + (3 \log 2) \cdot 2^r \\
    \Rightarrow \quad u^2/2  & > \frac{r-\ell}{(2^{r+1} - 2^{\ell})} + \frac{(3 \log 2) \cdot 2^{r}}{2^{r+1}- 2^{\ell}} ~.
\end{align*}
Note that the last term is a decreasing function of $r$, and the maximum is achieved at $r=\ell$ when we have
\begin{equation*}
    u^2/2 > (3 \log 2) \quad u > \sqrt{6 \log 2}~.
\end{equation*}
Thus, the bound holds for $u > u_0$ for a constant $u_0$. \qed

{\em Proof of Lemma~\ref{lemm:offd-lp-2nd}:} For $A \in \cA$ let $S = \{ A^T A \x : \x \in B_2^p \}$. Since $\bxi'$ satisfies (SP-1) and (SP-2), the random variable $\langle \bxi', A^T A \bxi \rangle$ is a weighted sum of a centered sub-Gaussian random variables when conditioned on $\bxi,F$. Then, we have 
\begin{align*}
    \| \langle A \bxi, A \bxi' \rangle \|_{L_p} & = \left( \E_{\bxi,\bxi'} |\langle A \bxi, A \bxi' \rangle|^p   \right)^{1/p} \\
    & = \left( \E_{\bxi,F} \left[ \E_{\bxi'|\bxi,F}  |\langle  \bxi', A^T A \bxi \rangle|^p  \right] \right)^{1/p} \\
    & \leq \left( \E_{\bxi,F} \left[ L^p \sqrt{p}^p \| A^T A \bxi \|_2^p \right] \right)^{1/p} \\
    & \leq L \sqrt{p} \left( \E_{\bxi} \left[ \sup_{y \in S} |\langle y, \bxi \rangle |^p \right] \right)^{1/p}~.
\end{align*}
Now, from Lemma~\ref{lemm:lpnorm}, we have 
\begin{equation*}
    \left( \E_{\bxi} \left[ \sup_{y \in S} |\langle y, \bxi \rangle |^p \right] \right)^{1/p} \leq 
    \E_{\bgg} \left[ \sup_{\y \in S} |\langle \bgg, \y \rangle| \right] + \sup_{\y \in S} (\E_{\bxi}|\langle \bxi, \y \rangle|^p)^{1/p}~.
\end{equation*}
For the first term, we have
\begin{align*}
 \E_{\bgg} \left[ \sup_{\y \in S} |\langle \bgg, \y \rangle| \right] = \E_{\bgg} \| A^T A \bgg \|_2 
    \leq (E \| A^T A  \bgg \|_2^2 )^{1/2} = \| A^T A \|_F \leq \| A \|_F \| A \|_{2 \rightarrow 2}~.
\end{align*}
For the second term,
\begin{align*}
    \sup_{y \in S} (\E_{\bxi} |\langle \y, \bxi \rangle|^p )^{1/p} 
    =     \sup_{z \in B_2^p} (\E_{\bxi} |\langle A^T Az, \bxi \rangle|^p )^{1/p}
    \leq L \sup_{z \in B_2^p} \sqrt{p} \| A^T A z \|_2 = L \sqrt{p} \| A \|_{2 \rightarrow 2}^2~.
\end{align*}
Plugging these bounds on the two terms back and taking supremum over $A \in \cA$ completes the proof. \qed 

{\em Proof of Theorem~\ref{theo:offd}:} Since $\bxi'$ is a decoupled tangent sequence to $\bxi$ adapted to $F$, we have
\begin{align*}
    \| B_{\cA`}(\bxi) \|_{L_p} &= \left\| \sup_{A \in \cA} \left| \sum_{\substack{j,k=1\\j \neq k}}^n \xi_j \xi_j \langle A_j, A_k \rangle \right| \right\|_{L_p} \\
    & \stackrel{(a)}{\leq} \left\| \sup_{A \in \cA} \left| \sum_{j,k=1}^n \xi_j \xi'_j \langle A_j, A_k \rangle \right| \right\|_{L_p} \\
    & \stackrel{(b)}{\leq} \gamma_2(\cA,\| \cdot \|_{2 \rightarrow 2}) \cdot \| N_{\cal A}(\bxi) \|_{L_p} + \sup_{A \in \cA} \| \langle A \bxi, A \bxi' \rangle \|_{L_p}\\
    & \stackrel{(c)}{\leq} \gamma_2(\cA,\| \cdot \|_{2 \rightarrow 2}) \cdot \big( \gamma_2(\cA,\| \cdot \|_{2 \rightarrow 2}) + d_F(\cA) \big) \\
    & \phantom{....} + \sqrt{p} \cdot d_{2 \rightarrow 2}(\cA) \cdot \big(\gamma_2(\cA,\| \cdot \|_{2 \rightarrow 2}) +  d_F(\cA) \big) + p \cdot d_{2 \rightarrow 2}^2(\cA)~,
\end{align*}
where (a) follows from Theorem~\ref{th:mds-dec}, (b) follows from Lemma~\ref{lemm:offd-lp}, and (c) follows from Lemma~\ref{lemm:scaling} and \ref{lemm:offd-lp-2nd}. That completes the proof. \qed

% \subsection{The Diagonal Terms: Bounded Random Variables}
% \label{ssec:diag}
% \input{sec/gen-diag-bnd.tex}

% \subsection{The Diagonal Terms: Unbounded sub-Gaussian Random Variables}
% \label{ssec:diag}
% \input{sec/gen-diag.tex}

\subsection{The Diagonal Terms}
\label{ssec:diag}
For the diagonal terms corresponding to (unbounded) sub-Gaussian random variables, we have the following main result:
\begin{theo}
Let $\cA \in \R^{m \times n}$ be a collection of $(m \times n)$ matrices. Let $\bxi = \{\xi_i\}$ be a stochastic process adapted to $F=\{F_i\}$ satisfying (SP-1) and (SP-2). Consider the random variable
\begin{equation}
    D_{\cA}(\bxi) = \sup_{A \in \cA} \left| \sum_{j=1}^n ( \xi_j^2 - E|\xi_j|^2 ) \| A^j \|_2^2 \right|~,
\end{equation}
where $A^j$ denotes the $j^{th}$ column of $A$. Then, we have 
% \begin{equation}
% \begin{split}
%     \| D_{\cA}(\xi) \|_{L_p} \leq &~ \gamma_{2 \rightarrow 2}(\cA, \| \cdot \|_{2 \rightarrow 2}) ~ [ \gamma_2(\cA, \| \cdot \|_{2 \rightarrow 2}) + d_F(\cA)] \\
%     & + \sqrt{p} \cdot d_{2 \rightarrow 2}(\cA) ~ [ \gamma_2(\cA, \| \cdot \|_{2 \rightarrow 2}) + d_F(\cA)] \\
%     & + p \cdot d_{2 \rightarrow 2}^2(\cA)~.
% \end{split}
% \end{equation}
\begin{equation}
\begin{split}
    \| D_{\cA}(\bxi) \|_{L_p} & \leq  
    \gamma_2(\cA, \| \cdot \|_{2 \rightarrow 2}) \big( \gamma_2(\cA, \| \cdot \|_{2 \rightarrow 2}) + d_F(\cA) \big) \\
    & \phantom{\leq} ~+ \sqrt{p} \cdot d_{2 \rightarrow 2}(\cA) \big( \gamma_2(\cA, \| \cdot \|_{2 \rightarrow 2}) + d_F(\cA) \big) + p \cdot d_{2 \rightarrow 2}^2(\cA) ~.
%    \| D_{\cA}(\xi) \|_{L_p} \leq &~ \sqrt{\log n} \cdot \gamma_2(\cA, \| \cdot \|_{F}) +  \sqrt{p} \cdot d_F(\cA) + p \cdot d_{2,\infty}(\cA)~.
\end{split}
\end{equation}
\label{theo:subg-diag-sharp}
\end{theo}
% %where (a) follows from Theorem~\ref{theo:exp-width} and (b) follows from the majorizing measures theorem [][]. Further, from Proposition~\ref{prop:sube2}, we have 
%
% %\ab{Note that the bound can be stated in the $(a + \sqrt{p} \cdot b + p \cdot p)$ form using the upper bound on $\| N_{\cA} \|_{L_p}$ from Lemma~\ref{lemm:scaling}.}
%
% \abcomment{In the above sharper result, we have managed to drop the multiplicative $\sqrt{\log n}$ term in the bound.}
The proof of Theorem~\ref{theo:subg-diag-sharp} relies on three key results, viz.~symmetrization, contraction, and de-symmetrization, generalized from the classical realm of i.i.d.~random variables~\cite{leta91} to stochastic processes $\bxi$ satifying (SP-1) and (SP-2).
%\abmargincomment{this is a wish list -- starting to work on these results}

% \begin{figure}[t]
% \centering
%  \includegraphics[width = 0.6 \textwidth]{figs/mds_ci_lcb.pdf}
%  \caption[]{Graphical Model 2 (GM2) schematic of decoupled MDS $Z =\{Z_i\}$ adapted to the stochastic process $\{ F_i\}$ so that $\E[Z_i \mid F_{1:(i-1)}] = 0$ a.s. The full joint distribution of $(F,Z)$ is given by \eqref{eq:genz}.
% Note that there is no restriction on the conditional distribution $F_i \mid (F_{1:(i-1)},Z_i)$, so that $F_i$ can have arbitrary dependence on $F_{1:(i-1)}$ and $Z_i$. Further, as shown in Proposition~\ref{prop:genz}, $\{Z_i\}$ is conditionally independent given $F$, i.e., $P(Z_{1:n} \mid F_{1:n}) = \prod_{i=1}^n P(Z_i \mid F_{1:n}) = \prod_{i=1}^n P(Z_i \mid F_{1:i})$. 
% In the schematic, while we show arrows only to one r.v., e.g., $F_{n-1} \rightarrow Z_n$, the MDS $Z_n|F_{1:(n-1)}$ implies (potential) dependence on the entire history $F_{1:(n-1)}$. Similarly, $F_n|F_{1:(n-1)},Z_n$ is illustrated only with arrows from $F_{n-1},Z_n$ to $F_n$. }
% \vspace*{-4mm}
% \label{fig:cilcb}
% \end{figure}

\subsubsection{Symmetrization for Stochastic Processes}
Let $\bxi = \{\xi_i\}$ be a stochastic process adapted to $F=\{F_i\}$ satisfying (SP-1) and (SP-2).
Let $\cG$ be a class of (bounded) functions.  Then we have the following symmetrization result:
%
%Let $Z_{1:i} = \{Z_1,\ldots,Z_n\}$ and similarly $Z'_{1:i} = \{Z'_1,\ldots,Z'_i \}$. With this notation, $Z = Z_{1:n}$ and $Z' = Z'_{1:n}$.
%
\begin{lemm}
Let $\bxi=\{\xi_i\}$ be a stochastic process adapted to $F=\{F_i\}$ satisfying (SP-1) and (SP-2). Let $E = \{\eps_i\}$ be a set of i.i.d.~Rademacher random variables. Let $\cG$ be a class of bounded functions. Then,
\begin{equation}
\E_{\bxi,F}\left[\sup_{g \in G} \left( \sum_{i=1}^n w_i \big(  g(\xi_i) - \E_{\bxi,F}[g(\xi_i)]  \big) \right) \right] \leq 2 \E_{\bxi,F,E} \left[ \sup_{g \in \cG} \sum_{i=1}^n w_i \eps_i g(\xi_i)  \right]~.
%+ E_{\bxi,\bxi',E} \left[ \sup_{g \in \cG} \sum_{i=1}^n w_i \eps_i g(\bxi'_i)  \right]~.
\end{equation}
\label{lemm:symm1}
\end{lemm}
\proof  Let $\bxi'=\{\xi'_i\}$ be a decouple tangent sequence (DTS) to $\bxi$ satisfying (SP-1) and (SP-2).
%
% `shadow' stochastic process such that
% \begin{equation}
% P(\bxi_i \mid F_{1:i}) = P(\bxi'_i \mid F_{1:i}) ~
% \label{eq:tgtcp}
% \end{equation}
% and they are conditionally independent, i.e., $\bxi_i \perp \bxi'_i \mid F_{1:i}$ or equivalently
% \begin{equation}
% P(\bxi_i,\bxi'_i \mid F_{1:i} ) = P(\bxi_i \mid F_{1:i}) P(\bxi_i \mid F_{1:i})~.
% \label{eq:tgtcp2}
% \end{equation}
% Note that we can always construct such a shadow $\bxi'$ following the graphical model for $(F,\bxi)$ (see Figure~\ref{fig:tgtcilcb}). Further, following the analysis in Proposition~\ref{prop:genz}, we have 
% \begin{equation}
% p(\bxi'_{1:n} \mid F_{1:n}) = \prod_{i=1}^n p(\bxi'_i \mid F_{1:n}) ~,
% \label{eq:tgtci}
% \end{equation}
% and for each $i=1,\ldots,n$, $\bxi'_i \perp F_{(i+1):n} \mid F_{1:i}$ so that $P(\bxi'_i \mid F_{1:n}) = P(\bxi'_i \mid F_{1:i})$.
%
We first focus on the expectation $\E_{\bxi,F}[g(\xi_i)]$. Note that
\begin{align*}
\E_{\bxi,F}[g(\xi_i)] &= \E_{\xi_i,F_{1:i}}[g(\xi_i)] \\
&= \E_{F_{1:i}}\left[\E_{\xi_i \mid F_{1:i}} \left[g(\xi_i) \right] \right] \\
& \overset{(a)}{=} \E_{F_{1:i}}\left[\E_{\xi'_i \mid F_{1:i}} \left[g(\xi'_i) \right] \right] \\
& \overset{(b)}{=} \E_{F_{1:n}} \left[\E_{\xi'_i \mid F_{1:n}} \left[g(\xi'_i) \right] \right] \\
& \overset{(c)}{=} \E_{F_{1:n}} \left[\E_{\xi'_{1:n} \mid F_{1:n}}\left[g(\xi'_i)\right] \right] \\
& = \E_F \left[\E_{\bxi'|F} \left[ g(\xi'_i) \right] \right]~, 
\end{align*}
where (a) follows since $\P(\xi'|F_{1:i}) = \P(\xi|F_{1:i})$ since $\bxi'$ is a DTS to $\bxi$, (b) follows since $\xi'_i \perp F_{i+1:n}|F_{1:i}$ by (SP-2), and (c) follows since $\xi'_i \perp \xi'_j |F_{1:n}$ for $j \neq i$ by (SP-2).

Then, by definition, 
\begin{align*}
\E_{F,\bxi} & \left[\sup_{g \in G} \left( \sum_{i=1}^n w_i \big( g(\xi_i) - \E_{F,\bxi}[g(\xi_i)]  \big) \right) \right] \\
&= \E_{F,\bxi} \left[\sup_{g \in G} \left( \sum_{i=1}^n w_i \big( g(\xi_i) - \E_F\left[ \E_{\bxi' \mid F}\left[g(\xi'_i) \right] \right]  \big) \right) \right] \\
&= \E_{F,\bxi} \left[\sup_{g \in G} \left( \sum_{i=1}^n w_i \bigg( g(\xi_i) - \E_F\left[ \E_{\bxi'}\left[g(\xi'_i) \mid F \right] \right]  \bigg) \right) \right] \\
&= \E_{F,\bxi} \left[\sup_{g \in G} \left( \sum_{i=1}^n w_i \bigg(  \E_F\left[ \E_{\bxi'}\left[g(\xi_i) - g(\xi'_i)  \mid F \right] \right]  \bigg) \right) \right] \\
& \overset{(a)}{=} \E_{F,\bxi} \left[\sup_{g \in G} \E_F\left[  \sum_{i=1}^n w_i  \E_{\bxi'}\left[ g(\xi_i) - g(\xi'_i) ~\bigg|~ F \right]   \right] \right] \\
& \overset{(b)}{\leq} \E_{F,\bxi} \left[\sup_{g \in G} \left( \sum_{i=1}^n w_i  \E_{\bxi'}\left[g(\xi_i) - g(\xi'_i) ~\bigg|~ F \right]  \right) \right] \\
& \overset{(c)}{=} \E_{F,\bxi} \left[\sup_{g \in G} \E_{\bxi'} \left[ \sum_{i=1}^n w_i \left( g(\xi_i) - g(\xi'_i) \right) ~\bigg|~ F \right]  \right] \\
& \overset{(d)}{\leq} \E_{F,\bxi} \left[ \E_{\bxi'} \left[ \sup_{g \in G}  \sum_{i=1}^n w_i  \left( g(\xi_i) - g(\xi'_i) \right) ~\bigg|~ F \right]  \right] \\
& = \E_{F,\bxi,\bxi'} \left[  \sup_{g \in G}  \sum_{i=1}^n w_i \left( g(\xi_i) - g(\xi'_i) \right)  \right]~,
\end{align*}
where (a) follows by linearity of expectation, (b) follows by Jensen's inequality on $F$, (c) follows by linearity of expectation, and (d) follows by Jensen's inequality on $\bxi'$.

Since $\bxi' = \{\xi'_i\}$ is a decoupled tangent sequence to $\bxi$, for any fixed realization $f_{1:n}$ of $F$, we have
\begin{align*}
\P(\xi_i \leq z_i | f_{1:i}) = \P(\xi'_i \leq z_i | f_{1:i}) \quad & \Rightarrow \quad  \P(\xi_i \leq z_i | f_{1:n}) = \P(\xi'_i \leq z_i | f_{1:n})~, ~\text{and} \\
\xi \perp \xi'_i | f_{1:i} \quad & \Rightarrow \quad \xi \perp \xi'_i | f_{1:n}~,
\end{align*}
by (SP-2).
As a result, conditioned on $f_{1:n}$, $\xi_i$ and $\xi'_i$  are conditionally independent and identically distributed so that
\begin{equation}
w_i (g(\xi_i) - g(\xi'_i)) \mid f_{1:n} \qquad \text{and} \qquad w_i \eps_i (g(\xi_i) - g(\xi'_i))  \mid f_{1:n}~
\label{eq:idrad}
\end{equation} 
are identically distributed. Hence,
\begin{align*}
\E_{F,\bxi,\bxi'} & \left[  \sup_{g \in G}  \sum_{i=1}^n w_i \left( g(\xi_i) - g(\xi'_i) \right) \right] \\
& = \E_{F} \left[  \E_{\bxi,\bxi'} \left[ \sup_{g \in G}  \sum_{i=1}^n w_i \left( g(\xi_i) - g(\xi'_i) \right) \bigg| F \right] \right] \\
&  \overset{(a)}{=} \E_{F} \left[  \E_{E,\bxi,\bxi'} \left[  \sup_{g \in G}  \sum_{i=1}^n w_i \eps_i \left( g(\xi_i) - g(\xi'_i) \right) \bigg| F \right] \right] \\
& \overset{(b)}{\leq} \E_{F} \left[  \E_{E,\bxi,\bxi'} \left[  \sup_{g \in G}  \sum_{i=1}^n w_i \eps_i g(\xi_i)  \bigg| F \right] \right] + \E_{F} \left[  \E_{\bxi,\bxi'} \left[  \sup_{g \in G}  \sum_{i=1}^n w_i \eps_i g(\xi'_i) \bigg| F \right]  \right] \\
& \overset{(c)}{=} \E_{F} \left[  \E_{E,\bxi} \left[  \sup_{g \in G}  \sum_{i=1}^n w_i \eps_i g(\xi_i)  \bigg| F \right] \right] + \E_{F} \left[  \E_{E,\bxi'} \left[  \sup_{g \in G}  \sum_{i=1}^n w_i \eps_i g(\xi'_i)  \bigg| F \right] \right] \\
& = \E_{E,F,\bxi} \left[   \sup_{g \in G}  \sum_{i=1}^n w_i \eps_i g(\xi_i)  \right] + \E_{E,F,\bxi'} \left[   \sup_{g \in G}  \sum_{i=1}^n w_i \eps_i g(\xi'_i)  \right] \\
& \overset{(d)}{=} 2 \E_{E,F,\bxi} \left[   \sup_{g \in G}  \sum_{i=1}^n w_i \eps_i g(\xi_i)  \right] 
\end{align*}
where (a) follows from \eqref{eq:idrad}, (b) follows by Jensen's inequality, (c) follows since conditioned on $F$ the first term does not depend on $\bxi'$ and the second term does not depend on $\bxi$, and (d) follows since $(E,F,\bxi)$ and $(E,F,\bxi')$ are identicaly distributed. That completes the proof. \qed

% \abcomment{In machine learning theory, one typically uses $w_i = \frac{1}{n}$, i.e., each element of $\w$ is $\frac{1}{n}$.}
% \swcomment{In general, should I think of the $w_i$ being adaptively chosen as well?}
% \ab{The `in expectation' result can be generalized to `with high probability' results in a straightforward manner if $g$ has bounded range, say Range$(g) = [0,1]$ as is considered in ML theory. We will try to do a more general resul down the line making minimal assumptions about $g$.}

% \abcomment{So, we dont need a high probability result explicitly, we just need the generalization below, which helps bound the $L_p$ norms}

For our analysis, we need a more general form of the symmetrization result:
\begin{lemm}
Let $\bxi$ be a stochastic process adapted to $F$ satisfying (SP-1) and (SP-2). Let $H: \R_+ \mapsto \R_+$ be a convex function and let $\w=[w_i] \in \R^n$ be a (constant) vector such that $H(\sup_{g \in \cG} |w_i g(\xi_i)|) < \infty$ for all $i$. Let $E = \{\eps_i\}$ be a collection of i.i.d.~Rademacher random variables. Then, we have
\begin{equation}
    \E_{\bxi,F}\left[ H\left( \sup_{g \in \cG} \left|  \sum_{i=1}^n w_i \big( g(\xi_i) - \E_{\bxi,F}[g(\xi_i)]   \big) \right| \right)  \right] 
    \leq  \E_{\bxi,F,E} \left[ H \left(  2 \sup_{g \in \cG} \left| \sum_{i=1}^n w_i \eps_i g(\xi_i) \right| \right) \right]~.
\end{equation}
\label{lemm:symm3}
\end{lemm}
The proof follows from that of Lemma~\ref{lemm:symm1} by simply noting that our use of Jensen's inequality with $\sup$ can be extended to include the convex function $H$ as well.

\subsubsection{De-symmetrization for Stochastic Processes}
We also need a de-symmetrization result for our analysis.
\begin{lemm}
%Let $\cG$ be a class of symmetric functions and let 
Let $\bxi$ be a stochastic process adapted to $F$ satisfying (SP-1) and (SP-2). Let $\w=[w_i] \in \R^n$ be a (constant) vector. Let $E = \{\eps_i\}$ be a collection of i.i.d.~Rademacher random variables. Then, we have
\begin{equation}
         \frac{1}{2} \E_{E,F,\bxi} \left[ \sup_{g \in \cG} \left( \sum_{i=1}^n w_i \eps_i (g(\xi_i) - \E_{F,\bxi}[g(\xi_i)] ) \right) \right]
         \leq 
         \E_{F,\bxi} \left[ \sup_{g \in \cG} \left(  \sum_{i=1}^n w_i \big( g(\xi_i) - \E_{F,\bxi}[g(\xi_i)]  \big) \right) \right] ~.
\end{equation}
\label{lemm:desym}
\end{lemm}
\proof Let $\bxi'$ be a decoupled tangent sequence to $\bxi$ satisfying (SP-1) and (SP-2). Following the analysis in Lemma~\ref{lemm:symm1} we have 
\begin{align*}
 \frac{1}{2} \E_{E,F,\bxi} & \left[ \sup_{g \in \cG} \left( \sum_{i=1}^n w_i \eps_i (g(\xi_i) - \E_{F,\bxi}[\xi_i] ) \right) \right] \\
& \leq  \frac{1}{2} \E_{E,F,\bxi,\bxi'} \left[ \sup_{g \in \cG} \sum_{i=1}^n w_i \eps_i (g(\xi_i) - g(\xi'_i)) \right]~\\
& =  \frac{1}{2} \E_{F} \left[ \E_{E,\bxi,\bxi'} \left[ \sup_{g \in \cG} \sum_{i=1}^n w_i \eps_i (g(\xi_i) - g(\xi'_i)) \bigg| F \right] \right]~\\    
& \overset{(a)}{=} \frac{1}{2} \E_F \left[ \E_{\bxi,\bxi'} \left[ \sup_{g \in \cG} \sum_{i=1}^n w_i (g(\xi_i) - g(\xi'_i)) \bigg| F \right] \right] \\
& \overset{(b)}{=} \frac{1}{2} \E_F \left[ \E_{\bxi,\bxi'} \left[ \sup_{g \in \cG} \sum_{i=1}^n w_i \big\{ (g(\xi_i) - \E_{F,\bxi}[g(\xi_i)]) - (g(\xi'_i)-\E_{F,\bxi'}[g(\xi'_i)]) \big\} \bigg| F \right] \right] \\
& \overset{(c)}{\leq}  \frac{1}{2} \E_F \left[ \E_{\bxi,\bxi'} \left[ \sup_{g \in \cG} \sum_{i=1}^n w_i (g(\xi_i) - \E_{F,\bxi}[g(\xi_i)]) \bigg| F \right] \right] \\
& \phantom{= ...} + \frac{1}{2}  \E_F \left[ \E_{\bxi,\bxi'} \left[ \sup_{g \in \cG} \sum_{i=1}^n w_i (g(\xi'_i)- \E_{F,\bxi'}[g(\xi'_i)]) \bigg| F \right] \right] \\
& \overset{(d)}{=}  \frac{1}{2} \E_F \left[ \E_{\bxi} \left[ \sup_{g \in \cG} \sum_{i=1}^n w_i (g(\xi_i) - \E_{F,\bxi}[g(\xi_i)]) \bigg| F \right] \right] \\
& \phantom{= ...} + \frac{1}{2}  \E_F \left[ \E_{\bxi'} \left[ \sup_{g \in \cG} \sum_{i=1}^n w_i (g(\xi'_i)- \E_{F,\bxi'}[g(\xi'_i)]) \bigg| F \right] \right] \\
& \overset{(e)}{=} \E_{F,\bxi} \left[ \sup_{g \in \cG} \left( \sum_{i=1}^n w_i (g(\xi_i) - \E_{F,\bxi}[g(\xi_i)]) \right) \right]~,
\end{align*}
where (a) follows since, as shown in the analysis of Lemma~\ref{lemm:symm1}, conditioned on a realization $f_{1:n}$ of $F$ 
\begin{equation*}
\sum_{i=1}^n w_i  \big( g(\xi_i) - g(\xi'_i) \big) \mid f_{1:n} \qquad \text{and} \qquad \sum_{i=1}^n w_i  \eps_i \big( g(\xi_i) - g(\xi'_i) \big) \mid f_{1:n}
\end{equation*}
are identically distributed, (b)  follows since $E_{F,\bxi}[\xi_i] = E_{F,\bxi'}[\xi'_i]$, (c) follows by Jensen's inequality, (d) follows since conditioned on $F$ the first term does not depend on $\bxi'$ and the second term does not depend on $\bxi$, and (e) follows since $(F,\bxi), (F',\bxi')$ are identically distributed. That completes the proof. \qed

For our analysis, we need a mildly more general form of the de-symmetrization result:
\begin{lemm}
%Let $\cG$ be a class of symmetric functions and let 
%Let $\w=[w_i] \in \R^n$ be a (constant) vector. Then, we have
Let $\bxi$ be a stochastic process adapted to $F$ satisfying (SP-1) and (SP-2). 
Let $H: \R_+ \mapsto \R_+$ be a convex function and let $\w=[w_i] \in \R^n$ be a (constant) vector such that $H(\sup_{g \in \cG} |w_i g(\xi_i)|) < \infty$ for all $i$.
Let $E = \{\eps_i\}$ be a collection of i.i.d.~Rademacher random variables. Then, we have
\begin{equation}
          \E_{E,F,\bxi} \left[ H \left( \frac{1}{2} \sup_{g \in \cG} \left| \sum_{i=1}^n w_i \epsilon_i (g(\xi_i) - \E_{F,\bxi}[g(\xi_i)] ) \right| \right) \right] 
         \leq 
          \E_{F,\bxi} \left[  H \left( \sup_{g \in \cG} \left|  \sum_{i=1}^n w_i \big( g(\xi_i) - E_{F,\bxi}[g(\xi_i)]  \big) \right| \right) \right]~.
\end{equation}
\label{lemm:desym2}
\end{lemm}
The proof follows from that of Lemma~\ref{lemm:desym} by noting that the application of Jensen's inequality can be extended to include the convex function $H$.

\subsubsection{Contraction for Stochastic Processes}
For the analysis, we will also need a variant of the following result from \cite[Lemma 4.6]{leta91}:
\begin{lemm}
Let $H : \R_+ \mapsto \R_+$ be convex. Let $\{\eta_i\}$ and $\{\gamma_i\}$ be two symmetric sequences of real valued random variables such that for some constant $K \geq 1$ and every $i$ and $t > 0$ we have
\begin{equation}
    P( | \eta_i | > t) \leq K P(|\gamma_i| > t )~.
\end{equation}
Then, for any finite sequence $\{ \x_i \}$ in a Banach space,
\begin{equation}
    \E \left[ H \left( \left\| \sum_i \eta_i \x_i \right\| \right) \right] \leq \E\left[ H\left( K \left\| \sum_i \gamma_i \x_i \right\| \right) \right]~.
\end{equation}
\label{lemm:cont}
\end{lemm}

% \subsubsection{Symmetrization and De-symmetrization for MDSs}
% \abcomment{this subsection can be dropped}
% \begin{lemm}
% Let $F  : \R_+ \mapsto \R_+$ be convex. Then, for any finite MDS $\{X_i\}$ in a Banach space such that $E F(\| X_i \| ) < \infty$ for every $i$, we have
% \begin{equation}
%       E \left[ F \left( \frac{1}{2} \| \sum_i \eps_i X_i \| \right) \right] \leq
%           E \left[ F \left( \| \sum_i X_i \| \right) \right] \leq
%              E \left[ F \left( 2 \| \sum_i \eps_i X_i \| \right) \right] ~.
% \end{equation}
% \end{lemm}

% The result would follow from C.5.1 and C.5.2. 

\subsubsection{Proof of Theorem~\ref{theo:subg-diag-sharp}}
The results on symmetrization, contraction, and de-symmetrization for stochastic processes satisfying (SP-1) and (SP-2) will now be used in the proof of Theorem~\ref{theo:subg-diag-sharp}, which follows a similar argument due to \cite{krmr14} for the i.i.d.~setting.
We will also need the following result specific to a set of i.i.d.~Gaussian random variables $\g = \{g_j\}$. The result was established in \cite{krmr14}. 
\begin{lemm}
Let $\g = \{g_j\}$ be a set of i.i.d.~Gausian random variables with $g_j \sim N(0,1)$. Then,
\begin{equation}
    \| C_{\cA}(\g) \|_{L_p} \leq \gamma_2(\cA, \| \cdot \|_{2 \rightarrow 2}) ( \gamma_2(\cA, \| \cdot \|_{2 \rightarrow 2}) + d_F(\cA) )  + \sqrt{p} d_{2 \rightarrow 2} d_F(\cA) + p d_{2 \rightarrow 2}^2(\cA)~.
\end{equation}
\label{lemm:cgauss}
\end{lemm}
Our proof of Theorem~\ref{theo:subg-diag-sharp} reduces the analysis for stochastic processes satisfying (SP-1) and (SP-2) to that for i.i.d.~Gaussian, which can use Lemma~\ref{lemm:cgauss}, and additional terms which can be suitably bounded. The reduction to the Gaussian case will utilize our results on  
symmetrization, contraction, and de-symmetrization.

{\em Proof of Theorem~\ref{theo:subg-diag-sharp}.} 
By definition of $D_{\cA}(\bxi)$ and from \ref{lemm:symm3} characterizing symmetrization of stochastic processes $\{ \bxi\}$ satisfying (SP-1) and (SP-2),  we have
\begin{align*}
    \| D_{\cA}(\xi) \|_{L_p} & = \left\| \sup_{A \in \cA} \left| \sum_{j=1}^n ( \xi_j^2 - E|\xi_j|^2 ) \| A^j \|_2^2 \right| \right\|_{L_p} 
     \leq 2 \left\| \sup_{A \in \cA} \left| \sum_{j=1}^n \eps_j |\xi_j|^2  \| A^j \|_2^2 \right| \right\|_{L_p}~,
\end{align*}
where $\{ \eps_j\}$ is a set of independent Rademacher variables independent of $\bxi$. Let $\{ g_j \}$ be a sequence of independent Gaussian random variables. By (SP-1), since $\xi_j|f_{1:j}$ is a $L$-sub-Gaussian random variable~\cite{vers18}, there is an absolute constant $c$ such that for all $t > 0$
\begin{align*}
    \P \left( |\xi_j|^2 \geq t L^2 \big| f_{1:j} \right) & \leq c \P(g_j^2 \geq t)~.
\end{align*}
By taking expectation over all such realizations $f_{1:j}$, we have
\begin{align*}
\E_{f_{1:j} \sim F_{1:j}} \left[ \P \left( |\xi_j|^2 \geq t L^2 \big| f_{1:j} \right) \right]  & \leq \E_{f_{1:j} \sim F_{1:j}} \left[ c \P(g_j^2 \geq t) \right]  \\
\Rightarrow \qquad \qquad \P \left( |\xi_j|^2 \geq t L^2 \right) & \leq c \P(g_j^2 \geq t)~.
\end{align*}
Now note that $\eta_j = \eps_j |\xi_j|^2$ and $\gamma_j = \eps_j |g_j/L|^2$ are both symmetric, and for all $t > 0$
\begin{align*}
       P( | \eta_j | > t) & \leq c P(|\gamma_j| > t)~,
\end{align*}
where the re-scaling in $\gamma_j$ has helped absorb the constant $L$. Then, from contraction of stochastic processes as in Lemma~\ref{lemm:cont}, we have\footnote{Recall that we are ignoring leading multiplicative constants which do not affect the order of the results.} 
\begin{equation}
\begin{split}
    \| D_{\cA}(\xi) \|_{L_p} & \leq \left\| \sup_{A \in \cA} \left| \sum_{j=1}^n \eps_j |\xi_j|^2  \| A^j \|_2^2 \right| \right\|_{L_p} \\
    & \leq \left\| \sup_{A \in \cA} \left| \sum_{j=1}^n \eps_j |g_j|^2  \| A^j \|_2^2 \right| \right\|_{L_p} \\
    & \overset{(a)}{\leq} \left\| \sup_{A \in \cA} \left| \sum_{j=1}^n \eps_j (|g_j|^2 - 1) \| A^j \|_2^2 \right| \right\|_{L_p}  + \left\| \sup_{A \in \cA} \left| \sum_{j=1}^n \eps_j  \| A^j \|_2^2 \right| \right\|_{L_p} \\
    & \overset{(b)}{\leq} 2 \left\| \sup_{A \in \cA} \left| \sum_{j=1}^n (|g_j|^2 - 1) \| A^j \|_2^2 \right| \right\|_{L_p}  + \left\| \sup_{A \in \cA} \left| \sum_{j=1}^n \eps_j  \| A^j \|_2^2 \right| \right\|_{L_p} \\ 
    & \leq 2 \left\| D_{\cA}(\g) \right\|_{L_p} + \left\| \sup_{A \in \cA} \left| \sum_{j=1}^n \eps_j  \| A^j \|_2^2 \right| \right\|_{L_p}~,
\end{split}
\label{eq:diag0}
\end{equation}
where (a) follows from Jensen's inequality and since $E|g_j|^2 = 1$, and (b) follows by de-symmetrization following Lemma~\ref{lemm:desym} and since the convex function here is 1-Lipschitz.

By triangle inequality, we have 
\begin{equation}
\begin{split}
    \left\| D_{\cA}(\g) \right\|_{L_p} & \leq \left\| C_{\cA}(\g) \right\|_{L_p} + \left\| B_{\cA}(\g) \right\|_{L_p}  \\
 &   \leq \gamma_2(\cA, \| \cdot \|_{2 \rightarrow 2}) ( \gamma_2(\cA, \| \cdot \|_{2 \rightarrow 2}) + d_F(\cA) ) \\
 & \phantom{\leq} + \sqrt{p} d_{2 \rightarrow 2}(\cA) (d_F(\cA) + \gamma_2(\cA, \| \cdot \|_{2 \rightarrow 2})) + p d_{2 \rightarrow 2}^2(\cA)~,
 \end{split}
 \label{eq:diag11}
\end{equation}
where we have used Lemma~\ref{lemm:cgauss} to bound $\left\| C_{\cA}(\g) \right\|_{L_p}$ and Theorem~\ref{theo:offd} to bound $\left\| B_{\cA}(\g) \right\|_{L_p}$. 

Further, note that $\sum_j \eps_j \| A^j \|_2^2$ is a sub-Gaussian stochastic process indexed over $A \in \cA$ relative to the metric
\begin{align*}
        d_2(A,B) &= ~ \left( \sum_{j=1}^n (\| A^j \|_2^2 - \| B^j \|_2^2 )^2 \right)^{1/2} \\ 
    & = ~ \left( \sum_{j=1}^n ( \| A^j \|_2 - \| B^j \|_2 )^2 \cdot (\| A^j \|_2 + \| B^j \|_2 )^2 \right)^{1/2} \\
    & \stackrel{(a)}{\leq} ~ \left( \sum_{j=1}^n \| A^j - B^j \|_2^2 \cdot (\| A^j \|_2 + \| B^j \|_2 )^2 \right)^{1/2} \\
    & \leq ~ 2 d_F(\cA) \| A - B \|_{2 \rightarrow 2}~,
\end{align*}
where (a) follows from triangle inequality. Then, following Lemma~\ref{lemm:lpnorm}, Proposition~\ref{prop:subg}, and the majorizing measure theorem~\cite{tala14}, we have
\begin{align}
    \left\| \sup_{A \in \cA} \left| \sum_j \eps_j \| A^j \|_2^2 \right| \right\|_{L_p}
    \leq d_F(\cA) \gamma_2(\cA, \| \cdot \|_{2 \rightarrow 2}) + \sqrt{p} d_F(\cA) d_{2 \rightarrow 2}(\cA)~.
    \label{eq:diag2}
\end{align}
Note that both the terms also appear in the bound for $\| D_{\cA}(\bgg) \|_{L_p}$ in \eqref{eq:diag11}. Putting \eqref{eq:diag0}--\eqref{eq:diag2} together we have 
\begin{equation}
\begin{split}
    \left\| D_{\cA}(\bxi) \right\|_{L_p} 
 &   \leq \gamma_2(\cA, \| \cdot \|_{2 \rightarrow 2}) ( \gamma_2(\cA, \| \cdot \|_{2 \rightarrow 2}) + d_F(\cA) )  \\
 & \phantom{\leq} + \sqrt{p} d_{2 \rightarrow 2}(\cA) (d_F(\cA) + \gamma_2(\cA, \| \cdot \|_{2 \rightarrow 2})) + p d_{2 \rightarrow 2}^2(\cA)~.
 \label{eq:diag3}
 \end{split}
\end{equation}
That completes the proof. \qed

% \section{Proofs of Theorem 1 and Theorem 2}
% \label{sec:proofs-main}
% \input{sec/proofs-main}

%\input{app/azuma.tex}

\end{document}